\documentclass[11pt]{article}

\usepackage[final]{acl}

\usepackage{times}
\usepackage{latexsym}

\usepackage[T1]{fontenc}

\usepackage[utf8]{inputenc}

\usepackage{microtype}

\usepackage{inconsolata}
\usepackage{arydshln}
\usepackage[normalem]{ulem}
\usepackage[most]{tcolorbox}
\newtcolorbox{promptbox}[2][]{
    enhanced,
    colback=gray!5,
    colframe=gray!50,
    boxrule=0.5pt,
    arc=3pt,
    fontupper=\ttfamily,
    title={#2},
    listing only,
    listing options={
        basicstyle=\small\ttfamily,
        breaklines=true,
        numbers=left,
        numberstyle=\tiny\color{gray},
    },
    #1
}
\usepackage{amsmath}
\usepackage{amssymb}
\usepackage{mathtools}
\usepackage{amsthm}
\usepackage{subcaption}
\usepackage{url}
\usepackage{booktabs} 
\usepackage{enumerate}
\usepackage{enumitem}
\usepackage[table]{xcolor}
\usepackage{multirow}
\usepackage{xspace}
\newcommand{\our}{\textsc{Miner}\xspace}

\usepackage{amsmath,amsfonts,bm}

\def\eqref#1{equation~\ref{#1}}

\def\1{\bm{1}}

\def\vo{{\bm{o}}}

\DeclareMathAlphabet{\mathsfit}{\encodingdefault}{\sfdefault}{m}{sl}
\SetMathAlphabet{\mathsfit}{bold}{\encodingdefault}{\sfdefault}{bx}{n}

\newcommand{\E}{\mathbb{E}}

\title{\our: Mining Intrinsic Mastery for Data-Efficient RL in Large Reasoning Models}

\author{Shuyang Jiang$^{\clubsuit,\diamondsuit}$, 
Yuhao Wang$^{\spadesuit}$,
{ \bf Ya Zhang\thanks{Corresponding Author}$^{\spadesuit,\diamondsuit}$},
{\bf Yanfeng Wang$^{\spadesuit,\diamondsuit}$},
{\bf Yu~Wang\footnotemark[1]$^{\spadesuit,\diamondsuit}$}
 \\
  $^{\clubsuit}$Fudan University \\
  $^{\spadesuit}$Shanghai Jiao Tong University \\
  $^{\diamondsuit}$Shanghai Artificial Intelligence Laboratory \\
  \texttt{shuyangjiang23@m.fudan.edu.cn} \\
  \texttt{\{colane,ya\_zhang,wangyanfeng622,yuwangsjtu\}@sjtu.edu.cn}
}

\begin{document}
\maketitle
\begin{abstract}
Current critic-free RL methods for large reasoning models suffer from severe inefficiency when training on positive homogeneous prompts (where all rollouts are correct), resulting in waste of rollouts due to zero advantage estimates. 
We introduce a radically simple yet powerful solution to \uline{M}ine \uline{in}trinsic mast\uline{er}y (\our), that repurposes the policy's intrinsic uncertainty as a self-supervised reward signal, with no external supervision, auxiliary models, or additional inference cost. 
Our method pioneers two key innovations: (1) a token-level focal credit assignment mechanism that dynamically amplifies gradients on critical uncertain tokens while suppressing overconfident ones, and (2) adaptive advantage calibration to seamlessly integrate intrinsic and verifiable rewards. 
Evaluated across six reasoning benchmarks on Qwen3-4B and Qwen3-8B base models, \our achieves state-of-the-art performance among the other four algorithms, yielding up to \textbf{4.58} absolute gains in Pass@1 and \textbf{6.66} gains in Pass@K compared to GRPO.
Comparison with other methods targeted at exploration enhancement further discloses the superiority of the two newly proposed innovations.
This demonstrates that latent uncertainty exploitation is both necessary and sufficient for efficient and scalable RL training of reasoning models.
Code is available at \url{https://github.com/pixas/Miner}.
\end{abstract}

\section{Introduction}

\begin{figure}[t]
    \centering
    \includegraphics[width=\linewidth]{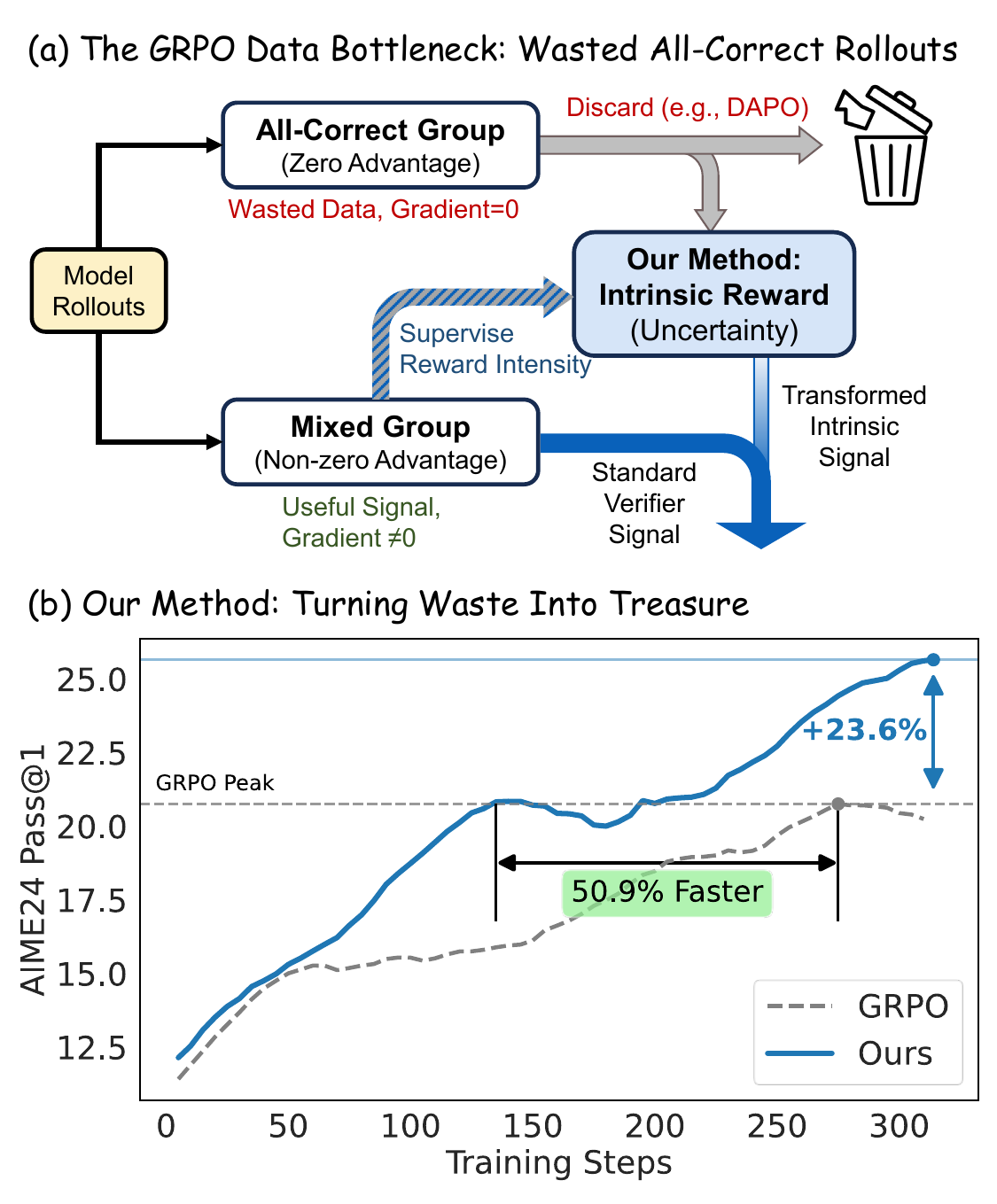}
    \caption{(a) Traditional GRPO algorithms produce a credible number of rollouts that do not contribute to RL updates, due to indistinguishable top rewards. (b) \our introduce intrinsic rewards to each rollout, injecting beneficial dense reward signals, achieving the same peak performance with only \textbf{50\%} training steps, and up to \textbf{23\%} higher performance on Qwen3-4B-Base.}
    \label{fig:intro}
\end{figure}

Reinforcement learning with verifiable rewards (RLVR) has become a central recipe for training large reasoning models (LRMs), enabling substantial reasoning gains from outcome-only supervision without relying on dense reward models or learned critics.
Critic-free algorithms such as GRPO~\citep{shao2024deepseekmath}, DAPO~\citep{yu2025DAPO}, and REINFORCE++~\citep{hu2025reinforce++} scale favorably by estimating advantages from multiple rollouts per prompt, making them particularly appealing for large-scale post-training.
Yet, this multi-rollout paradigm exposes an increasingly dominant inefficiency as base models strengthen: many prompts yield rollouts that all receive identical verifier rewards.
In these cases, GRPO-style relative advantage estimation collapses to (near-)zero, so the corresponding trajectories contribute no learning signal despite incurring full rollout cost~\citep{liu2025ghpo,sun2025improving,zhou2025april}.
Crucially, in the high-accuracy regime, \emph{positive homogeneous} (PH) prompts, where all sampled rollouts are correct, can occupy a large portion of each batch, rendering a non-trivial fraction of expensive rollouts computationally wasteful (see Fig.~\ref{fig:intro}(a)).

Existing efforts to mitigate homogeneity typically follow two paths, both of which introduce scaling trade-offs.
The first seeks to improve data quality through pre-filtering~\citep{xu2025not,yu2025DAPO,zheng2025act}, but these strategies inherently incur extra inference costs and their overhead grows as stronger models make more PH prompts.
Another line reuses past trajectories through rollout buffers~\citep{sun2025improving,jiang2025vcrl}, which introduces off-policy elements and distribution shifts that can complicate stability and large-scale deployment~\citep{xi2025bapo}.
This leaves a vital question unanswered: \emph{can we extract learning signals from PH prompts with essentially zero marginal overhead, while preserving the integrity of the primary verifier objective?}


In this work, we provide an affirmative answer by re-examining PH prompts through a simple yet underexploited lens: while PH rollouts are equally \emph{correct} according to the verifier, the underlying ``hard'' actions are not equally \emph{mastered} by the policy. 
Many correct solutions are generated via fragile, low-confidence reasoning paths that remain under-optimized if PH prompts are discarded. 
However, naively rewarding uncertainty is incompatible with the RLVR objective, as injecting intrinsic signals into heterogeneous prompts can blur the correctness boundary or overwhelm the outcome reward (Fig.~\ref{fig:cmp_exploratory}). 
To address this, we propose \our, a data-efficient framework that selectively targets PH prompts, transforming intrinsic uncertainty into a \emph{safe, bounded} learning signal.

\our comprises three tightly coupled designs:
(1) \textbf{Uncertainty-driven intrinsic rewards for PH only} where we define an intrinsic reward via per-token negative log-likelihood and apply \textbf{positive filtering} to exclusively reinforce under-confident but correct trajectories. This prevents the reinforcement of already-mastered modes and focuses optimization on the ``fragile'' reasoning paths.
(2) \textbf{Token-level focal credit assignment}, which is a focal reweighting mechanism that concentrates gradients on bottleneck tokens along the reasoning chain, using token probabilities as a discriminative weight to avoid uniformly reinforcing trivial tokens.
(3) \textbf{Adaptive advantage calibration} where we dynamically scale intrinsic advantages using a reference scale extracted from heterogeneous prompts within the same batch. This ensures a proper signal hierarchy, prioritizing the optimization of correctness while integrating intrinsic signals at an appropriate magnitude.
Notably, \our requires no additional rollouts, hints, replay buffers, or auxiliary reward models. By reusing quantities already computed during the PPO-style optimization process, it adds negligible overhead while reclaiming the utility of otherwise wasted rollouts.

We evaluate on two base models (Qwen3-4B-Base~\citep{qwen3technicalreport} and Qwen3-8B-Base) across six diverse reasoning benchmarks.
\our consistently outperforms GRPO variants and other strong baselines, achieving \textbf{+4.5} absolute Pass@1 on average and up to \textbf{$>10$} absolute gains in Pass@K on challenging benchmarks.
Further analyses of exploration dynamics, calibration stability, and cross-task transferability confirm that PH-targeted uncertainty exploitation is a general and robust strategy for enhancing reasoning models.
We summarize our contributions as follows:
\begin{enumerate}
    \item \textbf{Uncertainty-Driven Self-Supervised Reward:}
We introduce the first framework to transform a policy's intrinsic uncertainty into an informative reward signal for homogeneous prompts. By eliminating the need for external hints or auxiliary models, we unlock training signals for approximately 25\% of otherwise wasted rollouts at zero marginal cost.
    \item \textbf{Token-Focal Credit Assignment Mechanism:}
We propose a fine-grained focal weighting strategy that dynamically amplifies learning signals on critical, uncertain tokens while suppressing overconfident ones. This level of granularity overcomes sequence-level uniformity and prevents mode collapse, providing a precision entirely unexplored in existing RLVR literature.
    \item \textbf{State-of-the-Art Empirical Efficiency} Extensive experiments demonstrate that \our achieves significant gains in both sample efficiency and accuracy over all competitive baselines. With zero additional inference overhead, these results validate that latent uncertainty exploitation is a sufficient and necessary component for scalable RLVR training.
\end{enumerate}

\section{Preliminaries}
\label{sec:prelim}
\paragraph{Reinforcement learning with Verifiable Rewards (RLVR)}
The RL objective for the policy $\pi_{\theta}$ is to maximize the cumulative rewards $r$ received from the verifier. 
Specifically, Policy Gradient~\citep{williams1992simple} gives the following objective function:
\begin{equation}
\label{policy_gradient}
    \nabla \mathcal{J}_{\theta}=\E_{q\sim \mathcal{D},\vo\sim \pi_\theta(q)}\sum_{j=0}^T\nabla_\theta \log \pi_\theta (o_j\mid \vo_{<j}) A_j,
\end{equation}
where $\mathcal{D}$ is the training distribution, $q$ is an input prompt, $\vo$ is an output sequence consisting of $T$ tokens $\{o_1, o_2,\dots,o_T\}$, and $A_j$ is the advantage of the $j$-th token given the state $\vo_{<j}$.
Recently, DeepSeek-R1~\citep{guo2025deepseek} boosted large language models' reasoning ability via the Group Relative Policy Optimization~(GRPO; \citet{shao2024deepseekmath}) algorithm.
Each rollout is labeled by a verifiable reward $r(\cdot)$ which assigns 1 for correctness and 0 otherwise, and its advantage is estimated using the group average and standard deviation values of rewards from a group of $G$ trajectories $\mathcal{O}=\{\vo_i\}_{i=1}^G$ generated based on the same $q$:

\begin{equation}
\label{grpo_adv}
    A_i=\frac{r(\vo_i)-\mathrm{mean}(r(\vo_1),\dots,r(\vo_G))}{\mathrm{std}(r(\vo_1),\dots,r(\vo_G))}.
\end{equation}
GRPO optimizes the policy using the PPO objective~\citep{schulman2017proximal}:
\begin{multline}
    \label{ppo_update}
            \mathcal{J}_{\rm GRPO}(\theta)=\mathbb{E}_{q\sim \mathcal{D},\{\vo_i\}_{i=1}^G\sim \pi_{\theta}(\cdot\mid q)}\Bigg[\sum_{j=1}^{\vert \vo_i\vert}\min\\ \left(\rho_{i,j}A_i, 
    \mathrm{clip}(\rho_{i,j},1-\epsilon,1+\epsilon)A_i\right)\Bigg]-\\\beta D_{\rm KL}(\pi_{\theta}\|\pi_{\rm ref}),
\end{multline}
where $\rho_{i,j}=\frac{\pi_{\theta}(o_{i,j}\mid o_{i,<j},q)}{\pi_{\mathrm{old}}(o_{i,j}\mid o_{i,<j},q)}$ is the importance sampling ratio, $\vert \vo_i\vert$ is the sequence length and KL divergence~\citep{kullback1951information} \( D_{\mathrm{KL}}(\pi_{\theta} \,\|\, \pi_{\mathrm{ref}}) \), serves as a regularizer that encourages the policy \( \pi_{\theta} \) to remain close to the reference policy \( \pi_{\mathrm{ref}} \) in distributional space.

\paragraph{Data Efficiency}
Under the definition of RLVR, we simplify the reward function $r(\cdot)$ as a binary indicator, i.e., its value equals 1 for correct rollouts and 0 otherwise.
Under this setting, we classify a prompt $q$ into three categories, i.e., \uline{p}ositive \uline{h}omogeneous (PH), \uline{n}egative \uline{h}omogeneous (NH), and \uline{he}terogeneous (HE), based on the correctness of its $G$ rollouts $\{\vo_i\}_{i=1}^G$:
\begin{align}
    q:=\begin{cases}
        q_{ph} & \text{if } \sum_{i=1}^G r(\vo_i)=G \\
        q_{he} & \text{if } \sum_{i=1}^G r(\vo_i)\in (0, G) \\
        q_{nh} & \text{if } \sum_{i=1}^G r(\vo_i)=0
    \end{cases}
\end{align}
Under this definition, it is observed that rollouts would receive zero advantage when generated from PH and NH groups.
Given the rapidly evolving rate of LLMs~\citep{kaplan2020scaling,xiao2025densing}, the PH groups increasingly dominate the training batch for future base models, thereby causing substantial useless rollout costs.

\paragraph{Addressing the Diminishing Advantage Issue}
Extensive research has explored how to mitigate NH prompts by reducing prompt difficulty, e.g., appending hints~\citep{liu2025ghpo}, incorporating in-context demonstrations~\citep{bamba2025xrpo}, or injecting replay buffers~\citep{sun2025improving,jiang2025vcrl}.
We view these NH-oriented techniques as largely complementary to our goal and thus focus on PH prompts in this work.
In contrast, despite the increasing prevalence of PH groups in training batches as base model capabilities rapidly improve~\citep{kaplan2020scaling,xiao2025densing}, strategies for efficiently leveraging positive homogeneous responses remain under-explored.
Existing attempts are often suboptimal: DAPO~\citep{yu2025DAPO} adopts over-sampling and filtering, which still wastes rollouts; other approaches introduce denser rewards via implicit process reward models~\citep{yuan2025free,fei2025self} or cooperation with strong reward models~\citep{tao2025hybrid}, but they typically require SFT-tuned models or incur substantial computational overhead, limiting their use in zero-RL and large-scale settings.
Therefore, we aim to transform PH responses into heterogeneous ones by leveraging uncertainty-based intrinsic rewards, without incurring additional rollouts or relying on large learned reward models.

\begin{figure*}[tbp]
    \centering
    \includegraphics[width=\linewidth]{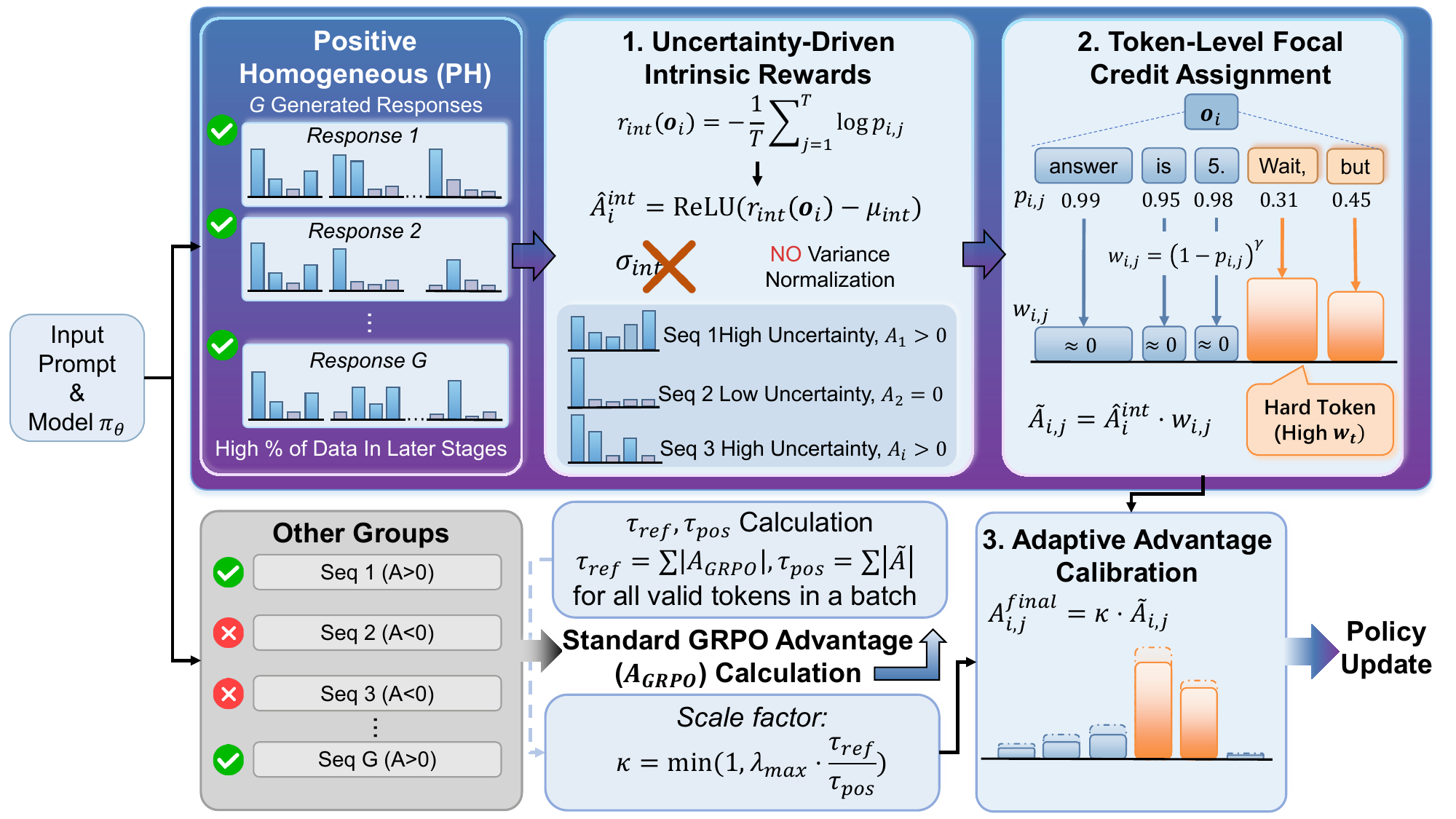}
    \caption{Framework of \our. We focus on introducing intrinsic rewards to positive homogeneous prompts (PH). \textbf{\textit{Upper Center}}: We use sequence-level uncertainty computed via the old policy $\pi_{\rm old}$ as the intrinsic rewards, to reinforce correct yet uncertain rollouts, without overfitting to already-mastered sequences; \textbf{\textit{Upper Right}}: Then, we leverage token-level focal credit assignment to specifically rewarding critical tokens, again skipping self-confident tokens; \textbf{\textit{Lower Right}}: Finally, to balance the learning signals from two groups, we calibrate the advantage score to a predefined threshold, significantly enhancing data efficiency without disturbing normal learning progress.}

    \label{fig:framework}

\end{figure*}

\section{\our}
\label{sec:method}

We introduce \our, a data-efficient RLVR framework that \emph{recovers training signals from Positive Homogeneous (PH) prompts}, to mine intrinsic mastery.
In standard GRPO, PH prompts yield zero advantage and are often filtered out, despite consuming the same rollout budget.
Our key hypothesis is that, although PH rollouts are \emph{equally correct} under the verifier, they are not equally \emph{mastered} by the policy.
\our converts the policy's intrinsic uncertainty into a dense \emph{mastery} signal, enabling online hard-positive mining to consolidate weak but correct reasoning modes.

\our consists of three components (Fig.~\ref{fig:framework}):
(i) \textbf{Uncertainty-Driven Intrinsic Rewards} to select \emph{which} correct trajectories remain weakly mastered (\S\ref{uncertainty_rewards});
(ii) \textbf{Token-Level Focal Credit Assignment} to localize \emph{where} the bottleneck steps are within a trajectory (\S\ref{focal_credit_assignment});
(iii) \textbf{Adaptive Advantage Calibration} to ensure the intrinsic mastery signal never overrides the extrinsic verification objective (\S\ref{adaptive_calibration}).

\subsection{Uncertainty-Driven Intrinsic Rewards}
\label{uncertainty_rewards}

For a PH prompt $q_{ph}$, all rollouts receive identical extrinsic reward $r(\vo_i)\equiv 1$, providing no learning direction.
We therefore define an intrinsic reward that reflects a lack of mastery.
Concretely, we use the per-token negative log-likelihood (NLL):
\begin{equation}
    r_{\rm int}(\vo_i)
    = -\frac{1}{T_i}\sum_{j=1}^{T_i}\log \pi_{\rm old}(o_{i,j}\mid o_{i,<j}, q).
\end{equation}
A correct response with a high NLL is treated as less mastered and thus more valuable to reinforce.
Importantly, we apply this intrinsic signal \emph{only} to PH prompts, avoiding interference with heterogeneous prompts where the verifier-defined reward is distinguishable.
We then compute a centered intrinsic advantage using a group-mean baseline:
\begin{equation}
\label{eq:miner_adv_init}
    A^{\rm int}_i
    = r_{\rm int}(\vo_i) - \frac{1}{G}\sum_{k=1}^G r_{\rm int}(\vo_k).
\end{equation}
Unlike GRPO, we omit standard-deviation normalization to preserve absolute uncertainty gaps (mild vs severe),
while relying on calibration (\S\ref{adaptive_calibration}) to control the global scale.
To avoid decreasing the probability of already well-mastered trajectories, we adopt \textbf{positive filtering}:
\begin{equation}
    \hat{A}^{\rm int}_i = \mathrm{ReLU}(A^{\rm int}_i),
\end{equation}
which only pulls up under-confident yet correct modes and circumvents penalizing confident and correct trajectories.

\begin{table*}[tbp]
  \centering
       \resizebox{\linewidth}{!}{%
    \begin{tabular}{lcccccccccccccc}
    \toprule
    \multirow{2}[4]{*}{\textbf{Model}} & \multicolumn{2}{c}{\textbf{AIME2024}} & \multicolumn{2}{c}{\textbf{AIME2025}} & \multicolumn{2}{c}{\textbf{AMC23}} & \multicolumn{2}{c}{\textbf{HMMT25}} & \multicolumn{2}{c}{\textbf{MATH}} & \multicolumn{2}{c}{\textbf{OlympiadB.}} & \multicolumn{2}{c}{\textbf{Avg.}} \\
\cmidrule(r){2-3} \cmidrule(r){4-5} \cmidrule(r){6-7} \cmidrule(r){8-9} \cmidrule(r){10-11} \cmidrule(r){12-13} \cmidrule(r){14-15}          & \textbf{P@1} & \textbf{P@K} & \textbf{P@1} & \textbf{P@K} & \textbf{P@1} & \textbf{P@K} & \textbf{P@1} & \textbf{P@K} & \textbf{P@1} & \textbf{P@K} & \textbf{P@1} & \textbf{P@K} & \textbf{P@1} & \textbf{P@K} \\
    \midrule
    \multicolumn{15}{c}{\textit{Base Model: Qwen3-4B-Base}} \\
    Base  & 9.51  & 50.00 & 6.65  & 60.00 & 32.91 & {97.50} & 0.99  & 26.67 & 58.89 & 91.00 & 31.81 & 65.40 & 23.46 & 65.10 \\
    GRPO  & 19.79 & 63.33 & 20.34 & \textbf{63.33} & 61.89 & {97.50} & 7.86  & 33.33 & 83.71 & 94.00 & 52.19 & 72.63 & 40.97 & 70.69 \\
    DAPO  & 21.22 & 53.33 & 18.70 & 50.00 & 64.51 & 95.00 & 6.72  & 30.00 & 82.11 & 94.40 & 51.17 & 71.26 & 40.74 & 65.67 \\
    R++   & 15.55 & 56.67 & 12.68 & 43.33 & 58.71 & 95.00 & 4.19  & 36.67 & 79.90 & 93.40 & 46.46 & 70.40 & 36.25 & 65.91 \\
    GSPO  & 16.04 & 70.00 & 11.98 & 46.67 & 58.09 & {97.50} & 4.53  & 30.00 & 80.25 & 93.40 & 46.21 & 69.88 & 36.38 & 67.91 \\
    UCAS & 19.09 & 56.67 & 15.42 & 53.33 & 60.14 & \textbf{100.00} & 9.19  & 36.67 & 82.80  & 94.00    & 49.46 & 70.05 & 39.35 & 68.45 \\
    \rowcolor[rgb]{ .867,  .922,  .969} \our  & \textbf{25.86} & \textbf{73.33} & \textbf{22.97} & 60.00 & \textbf{69.65} & {97.50} & \textbf{10.81} & \textbf{46.67} & \textbf{86.93} & \textbf{95.80} & \textbf{57.07} & \textbf{76.25} & \textbf{45.55} & \textbf{74.92} \\
    \midrule
    \multicolumn{15}{c}{\textit{Base Model: Qwen3-8B-Base}} \\
Base  & 9.17  & 53.33 & 5.91  & 40.00 & 39.14 & 95.00 & 1.51  & 23.33 & 56.31 & 91.40 & 34.51 & 66.09 & 24.43 & 61.53 \\
    GRPO  & 23.25 & 66.67 & 19.40 & 50.00 & \textbf{73.85} & 97.50 & 8.36  & 36.67 & 87.15 & 95.40 & 56.20 & 76.08 & 44.70 & 70.39 \\
    DAPO  & 25.26 & \textbf{70.00} & 18.02 & 60.00 & 67.79 & 97.50 & 12.32 & 46.67 & 87.74 & \textbf{98.20} & 56.42 & 76.94 & 44.59 & 74.89 \\
    R++   & 23.67 & \textbf{70.00} & 20.42 & 56.67 & 72.81 & 97.50 & 9.27  & 36.67 & 87.86 & 97.00 & 56.12 & 77.45 & 45.03 & 72.55 \\
    GSPO  & 25.00 & \textbf{70.00} & 19.68 & 53.33 & 70.27 & 95.00 & 9.16  & 46.67 & 87.56 & 95.40 & 56.21 & 75.90 & 44.65 & 72.72 \\
    UCAS & 22.16 & \textbf{70.00} & 15.34 & 53.33 & 66.00 & \textbf{100.00} & 8.88  & \textbf{50.00} & 84.54 & 94.80 & 53.36 & 74.87 & 41.71 & 73.83 \\
    \rowcolor[rgb]{ .867,  .922,  .969} \our  & \textbf{27.81} & \textbf{70.00} & \textbf{23.98} & \textbf{66.66} & 70.41 & \textbf{100.00} & \textbf{12.92} & \textbf{50.00} & \textbf{88.56} & 97.40 & \textbf{58.75} & \textbf{78.14} & \textbf{47.07} & \textbf{77.03} \\
    \bottomrule
    \end{tabular}%
    }
      \caption{Comprehensive comparison against other critic-free RL algorithms in terms of Pass@1 (P@1) and Pass@K (P@K) scores. ``OlympiadB.'' refers to the OlympiadBench. Best performance is highlighted with \textbf{bold}.}
  \label{tab:main_table}%
\end{table*}%

\subsection{Token-Level Focal Credit Assignment}
\label{focal_credit_assignment}

Sequence-level advantage assigns identical credit to all tokens, whereas uncertainty in reasoning is often concentrated at a few bottleneck steps.
We thus reweight token credits by a focal factor based on token probability
$p_{i,j} = \pi_{\rm old}(o_{i,j}\mid o_{i,<j}, q)$:
\begin{equation}
\label{eq:focal_weight}
    w_{i,j} = (1 - p_{i,j})^\gamma,
\end{equation}
where $\gamma \ge 0$ controls focusing ($\gamma{=}2$ suggested by \citet{lin2017focal}).
We treat $w_{i,j}$ as a constant (stop-gradient) to preserve the GRPO update form.
The token-level intrinsic advantage is:
\begin{equation}
    \tilde{A}_{i,j} = w_{i,j}\cdot \hat{A}^{\rm int}_i,
\end{equation}
which prioritizes bottleneck tokens and avoids spending gradient budget on trivial connectors.
\our introduces no additional rollouts and incurs negligible overhead, as it reuses token log-probabilities already computed by Eq.~(\ref{ppo_update}).

\subsection{Adaptive Advantage Calibration}
\label{adaptive_calibration}

Intrinsic mastery rewards and extrinsic verification rewards have different scales.
To respect the signal hierarchy, we cap the intrinsic advantage by a reference signal extracted from HE prompts in the same batch.
Let $\mathcal{B}_{he}$ and $\mathcal{B}_{ph}$ denote HE and PH prompts in the batch, respectively. We define
\begin{align}
    \tau_{\rm ref} &= \sum_{{q\in\mathcal{B}_{he},\, i\in\{1..G\}}} \sum_{j=1}^{\vert \vo_i^q \vert}
    \vert A_{i,j}^{q} \vert  \nonumber  \\  
    \tau_{\rm pos} &= \sum_{{q\in\mathcal{B}_{ph},\, i\in\{1..G\}}} \sum_{j=1}^{\vert \vo_i^q \vert}
    \vert \tilde{A}_{i,j}^{q} \vert,
\end{align}
and use a scale factor $\lambda_{\max}$ to guarantee that the intensity of additional advantages $\tau_{\rm pos}$ never exceed the configured signal threshold $\lambda_{\max}\cdot\tau_{\rm ref}$:
\begin{equation}
    \mathcal{A}^{\rm final}_{i,j} =
        \tilde{A}_{i,j}\cdot \min\left(1, \frac{\lambda_{\max}\cdot \tau_{\rm ref}}{\tau_{\rm pos}} \right)
\end{equation}
Finally, we optimize Eq.~(\ref{ppo_update}) using $\mathcal{A}^{\rm final}_{i,j}$ for PH prompts and standard GRPO advantages for others.

\section{Experiments}

\subsection{Experiment Setups}
\label{sec:exp_setup}
\paragraph{Evaluation} We adopt MATH500~\citep{lightman2023let}, AMC23~\citep{zwhe99_amc23}, OlympiadBench~\citep{he-etal-2024-olympiadbench}, AIME2024~\citep{aime2024} and AIME2025~\citep{aime2025}, HMMT25~\citep{balunovic_srimatharena_2025} as the evaluation testbeds with diverse complexity.
Apart from \textbf{GRPO}, we choose \textbf{DAPO}~\citep{yu2025DAPO}, \textbf{GSPO}~\citep{zheng2025group} and \textbf{REINFORCE++}~(R++; \citet{hu2025reinforce++}) as baselines.
To illustrate exploration superiority, we also include UCAS~\citep{xie2025unlocking} as a strong baseline.
We set the temperature as 0.7, top\_p as 0.95, and use a maximum token limit of 8192.
We conduct 128 rollouts for AIME2024, AIME2025, AMC23 and HMMT25, and 16 rollouts for OlympiadBench and MATH500.
We adopt Pass@1 and Pass@K~\citep{chen2021evaluating} as evaluation metrics to measure the exploitation and exploration abilities.


\paragraph{Training}
\label{sec:training_details}
We adopt DeepScaleR~\citep{deepscaler2025} as the training set and choose Qwen3-4B-Base~\citep{qwen3technicalreport} and Qwen3-8B-Base as the base policy.
We use veRL~\citep{sheng2024hybridflow} as the training framework.
We set $\lambda_{\max}$ to $1.5e-3$ for both models by grid search (elucidated in Appendix~\ref{sec:sensitivity_analysis}).
Additional hyperparameters for RL training are presented in Table~\ref{tab:hyperparameter}.


\subsection{Main Results}
As illustrated in Table~\ref{tab:main_table}, \our yields a \textbf{4.58} point increase in the Pass@1 metric and a consistent \textbf{4.23} point improvement in Pass@K on the Qwen3-4B model. These concurrent gains in both exploitation and exploration performance validate the comprehensive effectiveness of our approach. When applied to the more robust Qwen3-8B backbone, \our continues to effectively leverage PH problems. Compared to GRPO, the Pass@K improvement rises to 6.66, accompanied by a notable 2.37 gain in Pass@1. Notably, \our outperforms the strong DAPO baseline—even though DAPO benefits from oracle outcome verifier signals—achieving a +2.48 Pass@1 lead and a 2.14 Pass@K margin. This suggests that the intrinsic rewards for PH prompts are as informative as standard outcome rewards, highlighting a new path for scaling RLVR with diverse data. Furthermore, while hyperparameters were optimized on the 4B model, their seamless transfer to the 8B variant underscores \our's robustness and data efficiency when scaling to larger architectures.

\subsection{Ablation Study}
\label{sec:ablation}
In this section, we use Qwen3-4B to ablate our method with three variants: (1) \our without intrinsic reward~(\texttt{w/o IR}), which uses a fixed advantage score (0.05) for all positive homogeneous rollouts; (2) \our without focal weight~(\texttt{w/o FW}), which uses a uniform weight 1 for tokens throughout a trajectory; (3) \our without advantage calibration~(\texttt{w/o AC}), which allows for uncapped advantage signals.
Results in Fig.~\ref{fig:ablation_study} and Table~\ref{tab:ablation} demonstrate that rewarding positive homogeneous rollouts with a fixed, undistinguishable advantage is harmful for stable training, verifying that the improvements do come from the beneficial intrinsic rewards.
Moreover, focal weighting is extremely useful for improving models' exploration ability while simultaneously guaranteeing sharpening mastered knowledge, resulting in much higher Pass@K performance compared to the \texttt{w/o FW} variant.
Finally, without advantage calibration, the training fails in the middle and suffers from the under-fitting problem, which implies that a simple grid-search on $\lambda_{\max}$ is sufficient for \our to achieve fast convergence and stable training simultaneously.


\begin{table*}[tbp]
  \centering
           \resizebox{0.9\linewidth}{!}{%
    \begin{tabular}{lcccccccccccc}
    \toprule
    \multirow{2}[4]{*}{\textbf{Model}} & \multicolumn{2}{c}{\textbf{MedQA}} & \multicolumn{2}{c}{\textbf{MedMCQA}} & \multicolumn{2}{c}{\textbf{PubMedQA}} & \multicolumn{2}{c}{\textbf{MedXpertQA}} & \multicolumn{2}{c}{\textbf{MMLU-Pro}} & \multicolumn{2}{c}{\textbf{Avg}} \\
\cmidrule(r){2-3} \cmidrule(r){4-5} \cmidrule(r){6-7} \cmidrule(r){8-9} \cmidrule(r){10-11} \cmidrule(r){12-13}           & \textbf{P@1} & \textbf{P@K} & \textbf{P@1} & \textbf{P@K} & \textbf{P@1} & \textbf{P@K} & \textbf{P@1} & \textbf{P@K} & \textbf{P@1} & \textbf{P@K} & \textbf{P@1} & \textbf{P@K} \\
    \midrule
    Llama3.1-8B-Instruct  & 60.96 & 81.85 & 55.41 & 77.07 & 75.55 & 81.20 & 11.46 & 25.92 & 54.95 & 76.09 & 51.67 & 68.43 \\
    \quad +GRPO  & 70.15 & 84.45 & 63.64 & 76.12 & {77.20} & 82.00 & 16.29 & 29.35 & 65.57 & 77.39 & 58.57 & 69.86 \\
    \quad +HuatuoGPT-o1* & 70.20 & {-} & 58.20 & {-} & 76.10 & -     & 17.30 & {-} & 59.90 & {-} & 56.34 & {-} \\
    \quad +MedReason* & 68.40 & -     & 57.50 & -     & 77.60 & -     & 16.40 & -     & 63.10 & -     & 56.60 & - \\
    \rowcolor[rgb]{ .867,  .922,  .969} \quad +\our  & \textbf{72.03} & \textbf{86.17} & \textbf{64.59} & \textbf{78.29} & \textbf{78.95} & \textbf{82.40} & \textbf{17.53} & \textbf{32.69} & \textbf{66.82} & \textbf{78.96} & \textbf{59.96} & \textbf{71.70} \\
    \bottomrule
    \end{tabular}%
    }
  \caption{Comparison of \our with other data-centric methods, which outperforms the two representative medical models with consistent gains. ``*'' denotes the results are sourced from the original paper of \citet{wu2025medreasonelicitingfactualmedical}.}
  \label{tab:medical_cmp}%
\end{table*}%

\begin{figure}[tbp]
    \centering
    \includegraphics[width=\linewidth]{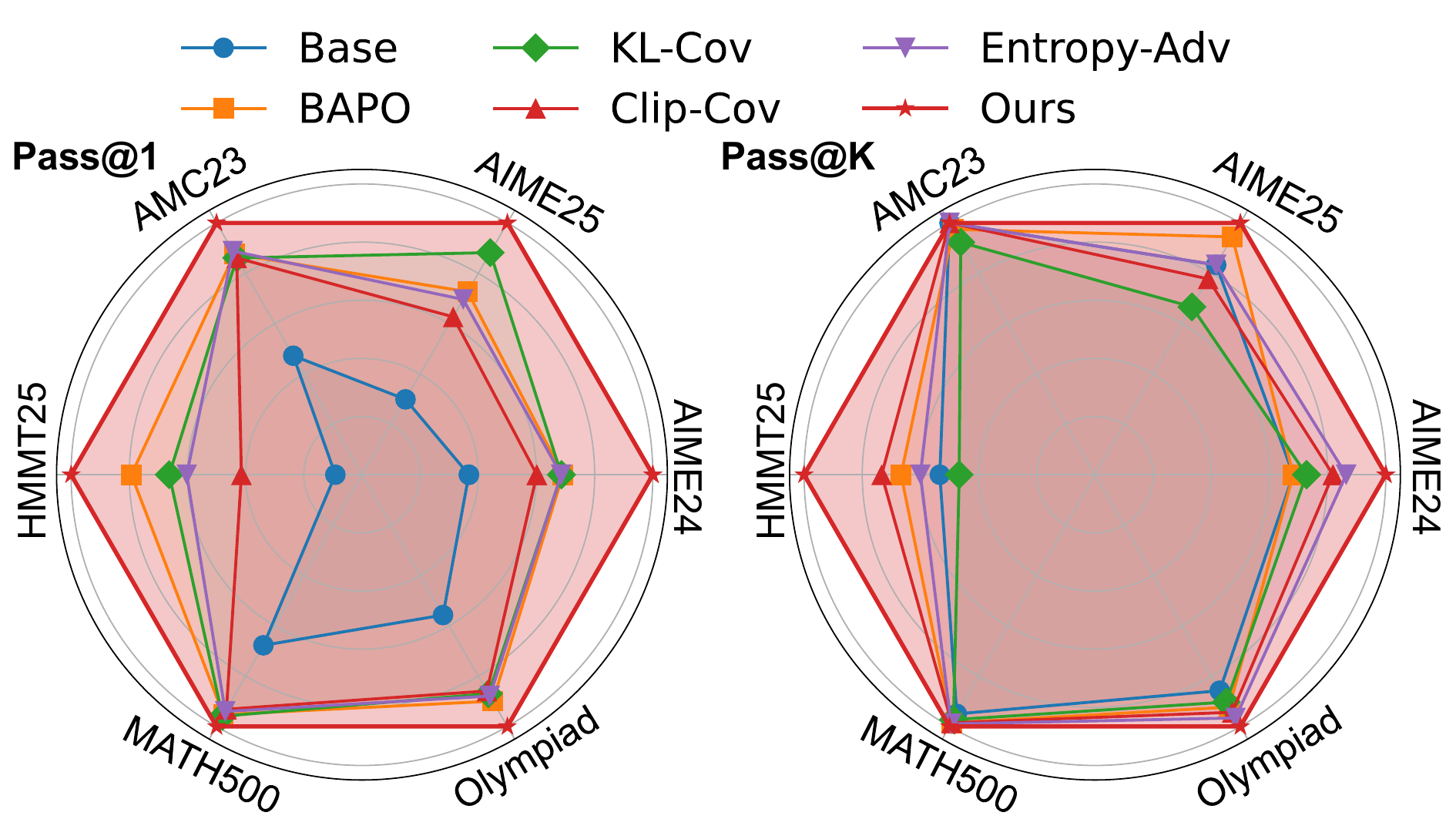}
    \caption{Comparison with other exploration-enhanced algorithms on normalized Pass@1 and Pass@K scores. 
    }

    \label{fig:cmp_exploratory}

\end{figure}

\section{Analysis}
In this section, we discuss the following research questions (RQ) of the \our algorithm:
\begin{enumerate}[parsep=0pt]
    \item [\textbf{RQ1:}] Can \our mitigate data inefficiency when altering the model backbone and domain task?
    \item [\textbf{RQ2:}] What is the relationship between \our and entropy-maximization methods?
    \item [\textbf{RQ3:}] Can \our benefit from larger token budgets?
    \item [\textbf{RQ4:}] How does \our conduct test-time scaling?
    \item [\textbf{RQ5:}] Can \our break the RL dilemma to incentivize beyond base capabilities?
    \item [\textbf{RQ6:}] Can \our stabilize training to further steps and larger base models?
\end{enumerate}

\begin{figure*}[t]
    \centering
    \begin{subfigure}{0.32\linewidth}
        \centering
        \includegraphics[width=\linewidth]{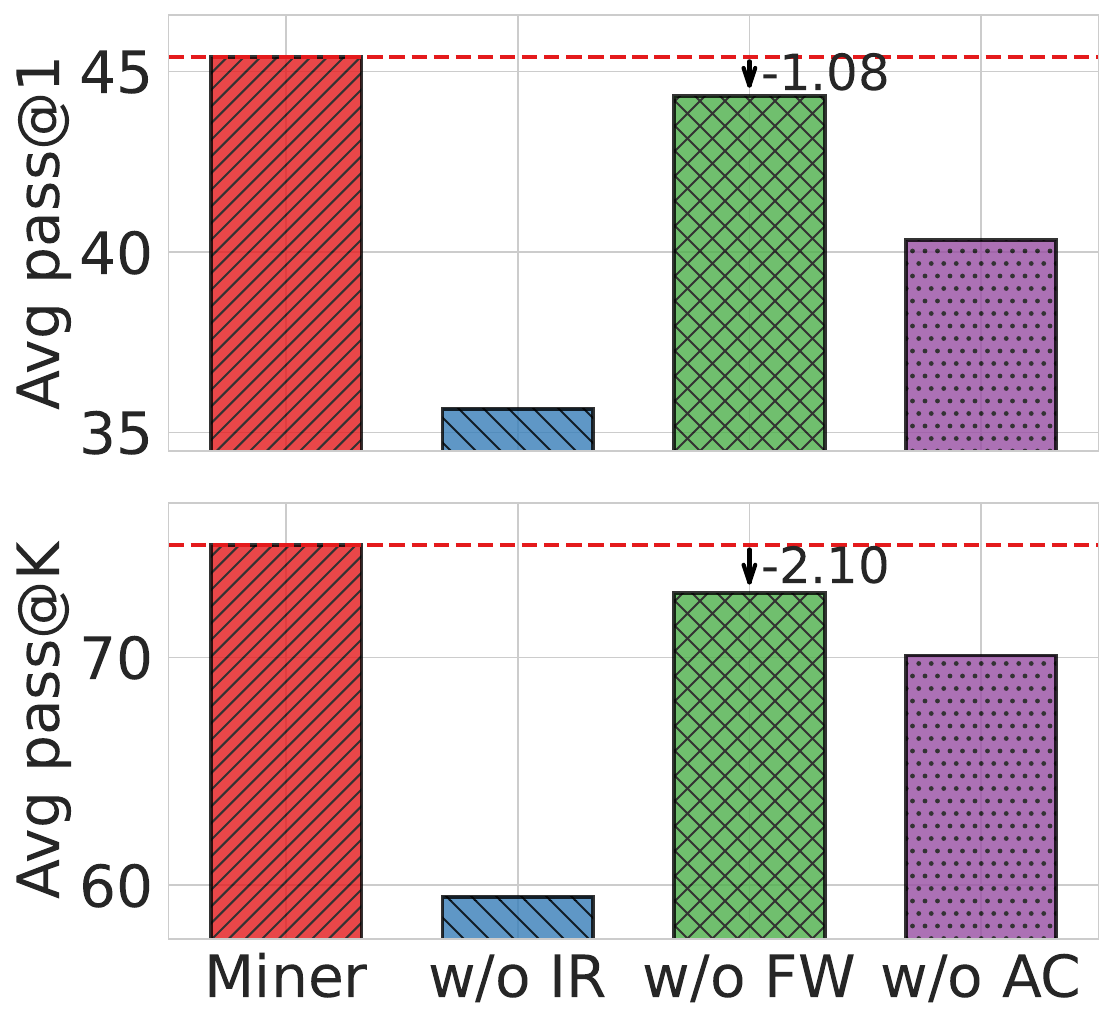}
        \caption{}
        \label{fig:ablation_study}
    \end{subfigure}
    \begin{subfigure}{0.32\linewidth}
        \centering
        \includegraphics[width=\linewidth]{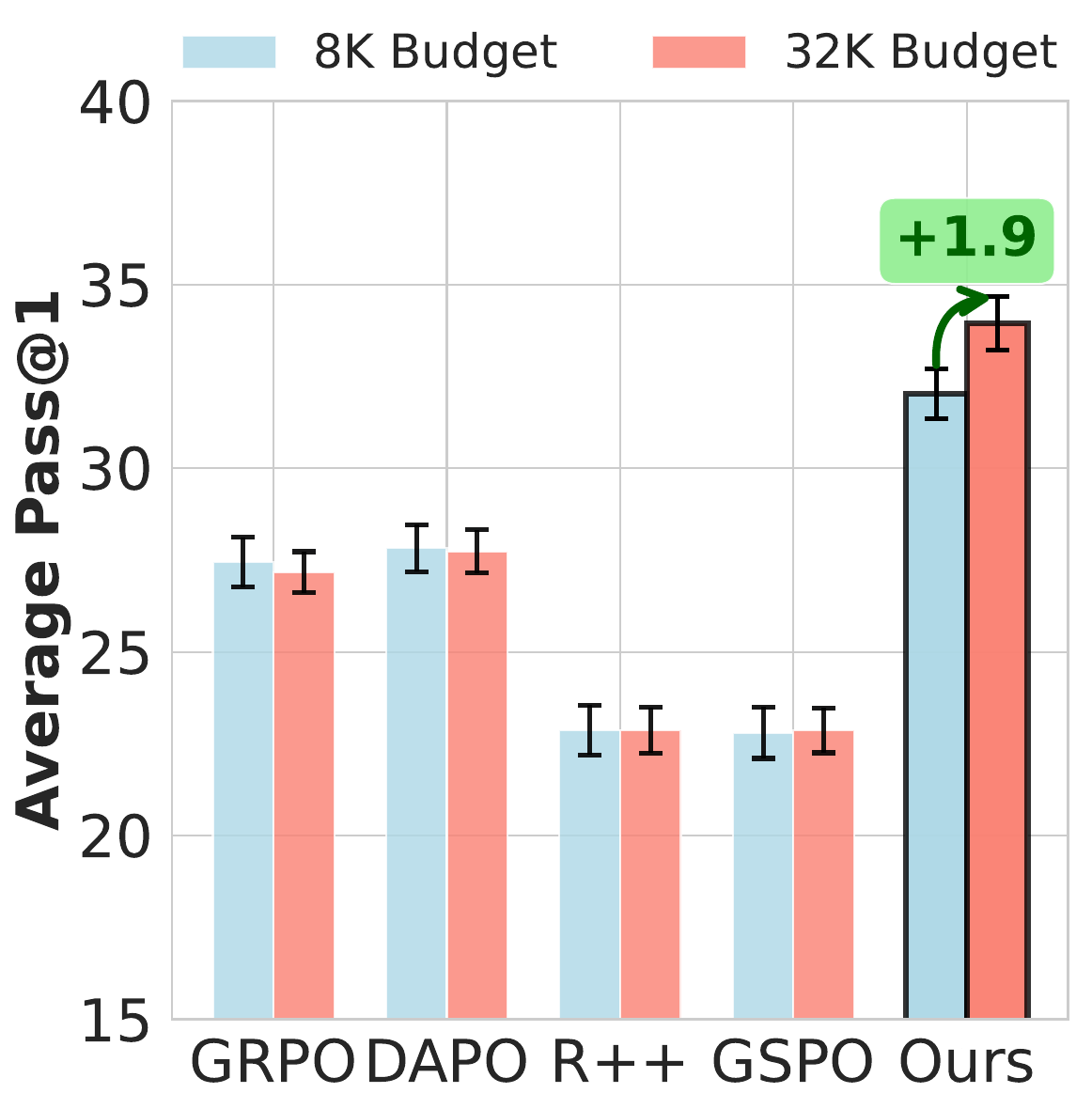}
        \caption{}
        \label{fig:scale_32K}
    \end{subfigure}%
    \begin{subfigure}{0.32\linewidth}
        \centering
        \includegraphics[width=\linewidth]{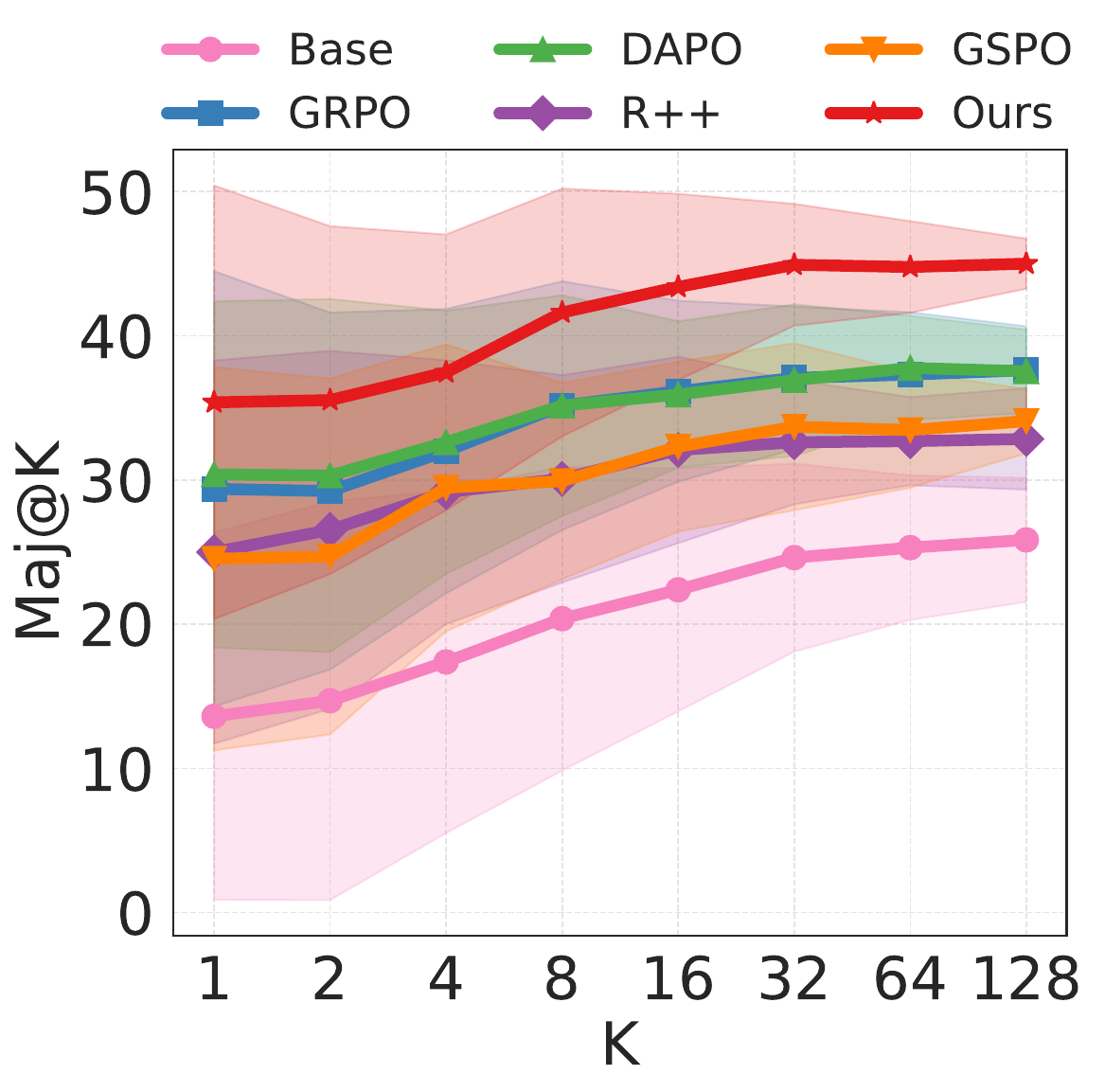}
        \caption{}
        \label{fig:maj_at_k}
    \end{subfigure}
    \caption{(a) Ablation study with three innovations (Intrinsic Reward (IR), Focal Weighting (FW) and Advantage Calibration (AC)) of \our on the Qwen3-4B base model. (b) Performance dynamics given sufficient inference budgets. Apart from fluctuation within the error bar, \our achieves non-negligible and sound improvements. (c) Parallel test-time scaling comparison with other algorithms, where \our consistently outperforms other baselines with over 5 absolute points. Shaded areas denote $\pm 1$ standard deviation over 10 runs.}
    \label{fig: analysis_experiments}

\end{figure*}

\paragraph{Response to RQ1: \our is domain- and backbone-agnostic, delivering consistent gains across diverse tasks and architectures.}
Given replication challenges for RL on Llama-family models in math reasoning (Appendix~\ref{sec:trial_on_llama}, \citep{gandhi2025cognitive,liu2025understanding}), we evaluate \our on medical reasoning, which is a practical domain feasible for scalable RL training. 
We test on MedQA~\citep{jin2021disease}, MedMCQA~\citep{pal2022medmcqa}, PubMedQA~\citep{jin-etal-2019-pubmedqa}, MedXpertQA~\citep{zuo2025medxpertqa}, and MMLU-Pro medical subsets~\citep{wang2024mmlupro}, using an 85K training corpus sub-sampled from MedQA/MedMCQA training partitions (strictly held out; Appendix~\ref{sec:details_of_med_corpus}). With 4 samples per question, statistical variance is minimized across these large-scale benchmarks.
To verify data efficiency, we compare against medical-specialized models HuatuoGPT-o1~\citep{chen2024huatuogpt} and MedReason~\citep{wu2025medreasonelicitingfactualmedical}, which require complex data processing and GPT-4o distillation despite sharing the Llama3.1-8B-Instruct backbone~\citep{DBLP:journals/corr/abs-2407-21783}. Table~\ref{tab:medical_cmp} shows \our surpasses GRPO and both data-intensive baselines across all five medical tasks of varying difficulty. Crucially, training logs (Fig.~\ref{fig:med_log}) confirm \our achieves stable improvement even with $>$65\% PH prompts per batch.
This demonstrates \our’s dual generalization capability: consistent performance gains across model architectures (backbone-agnostic) and domains (domain-agnostic), particularly valuable in data-scarce fields like medicine.

\begin{figure*}[t]
    \centering

    \includegraphics[width=\linewidth]{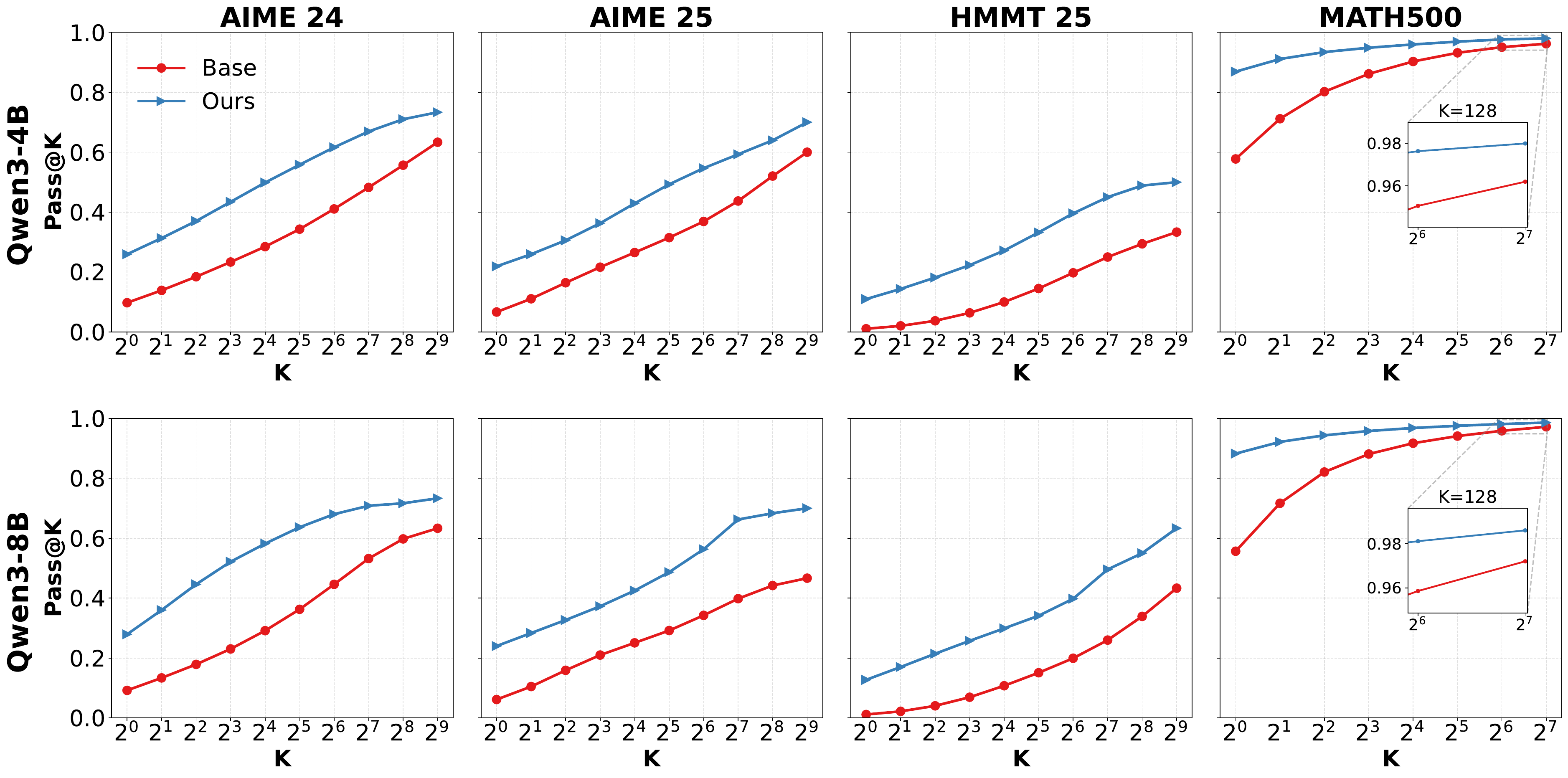}

    \caption{Pass@K scaling of \our and Base model on Qwen3-4B (\textbf{\textit{Upper}}) and Qwen3-8B (\textbf{\textit{Lower}}) models, where \our still demonstrates improvements for a sufficiently large $K$.}
    \label{fig: passk_scale}

\end{figure*}

\paragraph{Response to RQ2: \our improves the exploration via a healthier manner.}
We additionally compare \our with other exploration-enhancement methods that manipulate advantage signals on heterogeneous prompts, to unveil that operating on positive homogeneous prompts results in healthier improvement on exploration.
We build on Qwen3-4B and compare with BAPO~\citep{xi2025bapo}, KL-Cov and Clip-Cov~\citep{cui2025Entropya}, which enhance exploration by softening the upper clipping bound, as well as Entropy-Adv~\citep{cheng2025reasoning}, which explicitly rewards high-entropy tokens by shaping the advantage estimation with the actor entropy value.
We do not compare with entropy regularization, which will result in unbounded entropy collapse for inappropriate setups of the hyperparameter $\alpha$~\citep{cui2025Entropya}.
We use the suggested hyperparameters released at their official codebase~(details in Appendix~\ref{sec:details_of_other_exploration}).
We show the normalized performance in Fig.~\ref{fig:cmp_exploratory} and full results in Table~\ref{tab:cmp_other_entropy}, where these methods could not generalize as perfectly as in their original paper on both metrics.
With only one hyperparameter and stable control of shaped advantage estimations, \our surpasses them by a large margin.

\paragraph{Response to RQ3: Yes. \our improves with a larger inference budget.}
To validate whether the intrinsic rewards on positive homogeneous groups would both improve the model's performance under a less-constrained token limit~\citep{snell2025scaling,muennighoff2025s1}, we evaluate by extending the inference budget to the maximum context limit of 32K.
We choose Qwen3-4B as the base backbone and test on four challenging datasets (AIME24, AIME25, AMC23, and HMMT25), as they pose high demands for testing budget~\citep{guo2025deepseek}.
We plot the comparison in Fig.~\ref{fig:scale_32K}.
We observe that the performance of other methods fluctuates within the margin of error, demonstrating that they fail to achieve consistent gains from increased test-time compute budgets.
In contrast, our method exhibits a stable and statistically significant improvement (+1.9 pass@1), indicating that our algorithm not only enhances data efficiency during training but also substantially boosts the model's scaling potentials during deployment. 

\paragraph{Response to RQ4: \our could consistently improve with more parallel test-time compute.}
To illustrate the test-time scaling~\cite{zhang2025survey} potential of \our against other baselines, we adopt self-consistency~(SC; \citet{wang2023selfconsistency}) as the evaluation method under multiple parallel samples.
We do not compare sequential scaling as it is empirically verified to be highly inefficient compared to the parallel scaling paradigm~\citep{ghosal2025Does}.
Fig.~\ref{fig:maj_at_k} demonstrates the substantial scaling potentials of \our compared to other algorithms with 10 repetitions.
Most methods suffer from performance staleness, while \our improves consistently given more samples, surpassing other baselines by \textbf{7.38} points.
Detailed figures are presented at Table~[\ref{tab:aime24_majk_table},\ref{tab:aime25_majk_table},\ref{tab:amc23_majk_table},\ref{tab:hmmt25_majk_table}].

\paragraph{Response to RQ5: When sampling up to 512 trajectories, the answer is `yes'.}
\citet{yue2025does} reveals that previous RLVR algorithms, e.g., GRPO or DAPO, primarily sharpen the policy distribution but often sacrifice the potential for discovering optimal solutions when given an ample number of trials. 
In addressing this research question, we rigorously investigate whether \our can effectively mitigate the mode collapse issue under a sampling regime of 512 trajectories, a threshold considered sufficiently large to comprehensively unveil the model's underlying reasoning capabilities and behavioral patterns. 
To ensure a thorough evaluation, we test the models on highly challenging benchmarks such as AIME24, AIME25, and HMMT25; additionally, we include the comparatively accessible MATH500 dataset with $ K=128$ to provide a balanced assessment across difficulty levels. 
As illustrated in Fig.~\ref{fig: passk_scale}, \our consistently exceeds the base model's performance even at large $K$ values, with results derived from both Qwen3-4B and Qwen3-8B architectures. 
Notably, our approach exhibits a consistent and robust growth trajectory on demanding benchmarks like AIME25 and HMMT25, with no signs of performance plateauing or saturation, thereby indicating sustained exploratory capacity. 
In essence, \our substantially alleviates the mode collapse inherent in standard GRPO, achieving an optimal equilibrium between exploration and exploitation that enhances overall solution diversity and reliability.

\begin{table*}[tbp]
  \centering
  \caption{Scaling of \our to a larger Qwen3-14B-Base model, with consistent improvement against GRPO.}
  \resizebox{\linewidth}{!}{%
    \begin{tabular}{lcccccccccccccc}
    \toprule
    \multirow{2}[4]{*}{\textbf{Model}} & \multicolumn{2}{c}{\textbf{AIME2024}} & \multicolumn{2}{c}{\textbf{AIME2025}} & \multicolumn{2}{c}{\textbf{AMC23}} & \multicolumn{2}{c}{\textbf{HMMT25}} & \multicolumn{2}{c}{\textbf{MATH}} & \multicolumn{2}{c}{\textbf{OlympiadB}} & \multicolumn{2}{c}{\textbf{Avg}} \\
\cmidrule(r){2-3} \cmidrule(r){4-5} \cmidrule(r){6-7} \cmidrule(r){8-9} \cmidrule(r){10-11} \cmidrule(r){12-13} \cmidrule(r){14-15}         & \textbf{P@1} & \textbf{P@K} & \textbf{P@1} & \textbf{P@K} & \textbf{P@1} & \textbf{P@K} & \textbf{P@1} & \textbf{P@K} & \textbf{P@1} & \textbf{P@K} & \textbf{P@1} & \textbf{P@K} & \textbf{P@1} & \textbf{P@K} \\
    \midrule
    Qwen3-14B-Base & 9.77  & 53.33 & 5.55  & 43.33 & 37.85 & 97.50 & 1.95  & 26.67 & 53.22 & 91.20 & 34.37 & 68.16 & 23.79 & 63.37 \\
    \quad +GRPO & 23.15 & 70.00 & 20.65 & 60.00 & 72.48 & 97.50 & 9.32  & 40.00 & 87.90 & 95.80 & 55.49 & 76.59 & 44.83 & 73.32 \\
    \rowcolor[rgb]{ .867,  .922,  .969} \quad +\our & \textbf{32.71} & \textbf{73.33} & \textbf{24.90} & \textbf{66.67} & \textbf{73.69} & \textbf{100.00} & \textbf{13.85} & \textbf{53.33} & \textbf{88.90} & \textbf{98.00} & \textbf{59.56} & \textbf{79.69} & \textbf{48.94} & \textbf{78.50} \\
    \bottomrule
    \end{tabular}%
    }
  \label{tab:scale_to14}%
\end{table*}%

\begin{figure*}[tbp]
    \centering
    \includegraphics[width=\linewidth]{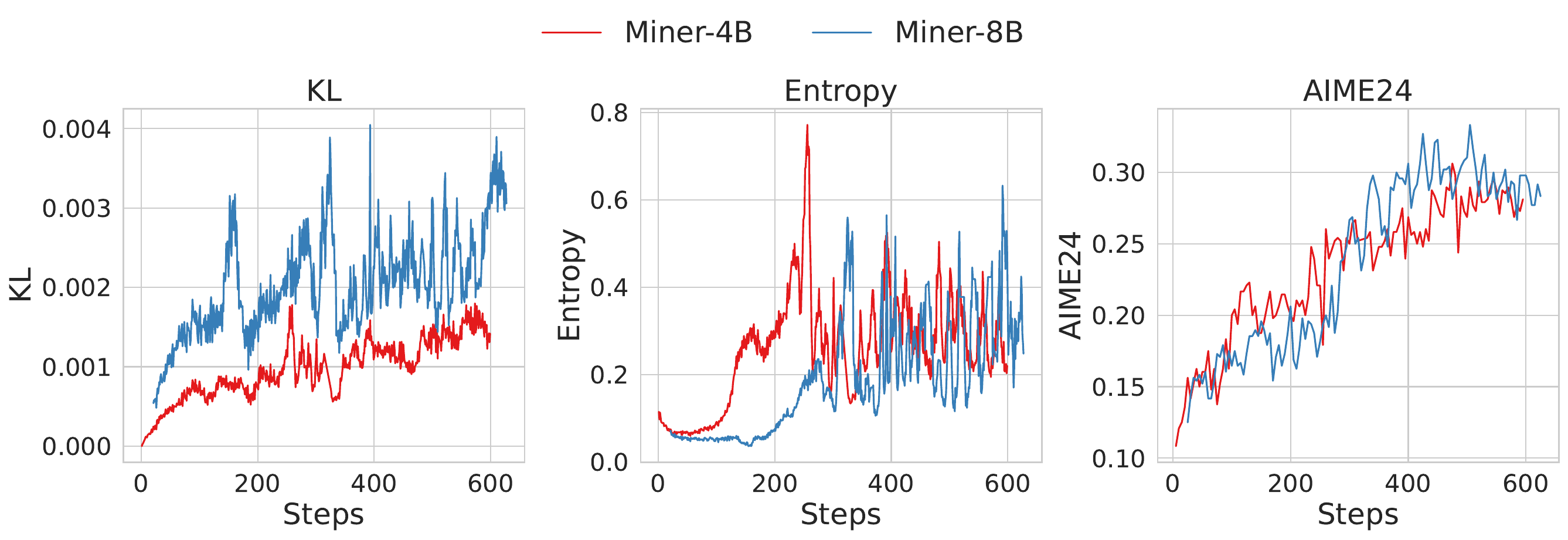}
    \caption{KL, Entropy and AIME24 scores when extending RL training to another 1 epoch on Qwen3-4B and Qwen3-8B backbones.}

    \label{fig:extend_training}

\end{figure*}

\paragraph{Response to RQ6: Yes, \our is consistently stable.}
To validate the stability of \our, we first scale it to Qwen3-14B-Base with the same $\lambda_{\max}$ value and compare it with GRPO.
Experiment results in Table~\ref{tab:scale_to14} demonstrate that \our is still superior to GRPO, where the improvements on Pass@1 and Pass@K remain consistent at 4.11 and 5.18, respectively.
This again validates the generality of our method and no requirements for dedicated hyperparameter tuning.
Besides capacity scaling, we also extend the training process to another training epoch.
We record stability metrics, including KL and entropy, and pass@1 score on the selected dev set, AIME24 and plot the dynamics in Fig.~\ref{fig:extend_training}.
Results reveal that although in the initial epoch the training dynamics fluctuate, both KL and entropy remain stable and the AIME24 score is consistently increasing, which cross-verifies the generalization of \our and its adaptability in prolonged RL training.

\section{Conclusion}
We present \our, a novel reinforcement learning framework that transforms previously wasted positive homogeneous (PH) rollouts into valuable learning signals through uncertainty-driven intrinsic rewards. 
By introducing sequence-level uncertainty rewards with positive filtering, token-level focal credit assignment, and adaptive advantage calibration, our method effectively converts gradient-desert PH groups into catalysts for knowledge consolidation. 
Extensive experiments demonstrate that \our boosts both pass@1 and pass@K, without additional computation or excessive use of hyperparameters. 
By turning ``solved'' prompts into robustness incubators, \our paves the way for efficient RL training where every rollout counts.

\section*{Limitations}
This work focuses on unlocking the learning signal from positive homogeneous (PH) prompts, and we validate \our across three model scales (Qwen3-4B, Qwen3-8B and Qwen3-14B) and a diverse suite of reasoning benchmarks. 
We did not further scale training to substantially larger backbones (e.g., 32B) due to computational constraints. 
Nevertheless, \our introduces negligible additional overhead and only a single new hyperparameter, $\lambda_{\max}$; importantly, the same setting transfers smoothly from 4B to 8B in our experiments, suggesting that scaling primarily requires additional compute rather than methodological changes.


\section*{Ethical Considerations}
Reinforcement learning from verifiable rewards has become a major part of bootstrapping large language models' intelligence in the data-scarce world.
While effective, a large proportion of rollout data is useless for training, which brings severe training inefficiency.
Our work aims to offer a practical solution to resolve the data inefficiency problem via a robust and computationally friendly manner, fostering innovation and collaboration to accelerate advancements that ultimately benefit society.

\section*{Acknowledgments}
This work was supported by the National Natural Science Foundation of China (No. 62576209) and STCSM (No. 2025SHZDZX025G05).

\bibliography{custom}

\appendix


\begin{table*}[tbp]
  \centering
  \caption{Hyperparameters for \our training.}
    \begin{tabular}{lcc}
    \toprule
    \textbf{Hyperparameter} & \multicolumn{1}{c}{\textbf{Qwen3-4B/8B-Base}} & \multicolumn{1}{c}{\textbf{Llama3.1-8B-Instruct}} \\
    \midrule
    max response length & 8192 & 8192 \\
    batch szie & 128   & 128 \\
    rollout batch size & 128 & 128 \\
    learning rate & 2.0e-06 & 1.0e-06 \\
    total training epochs & 1     & 1 \\
    rollout number & 16    & 16 \\
    PPO clip range $\epsilon$ & 0.2   & 0.2 \\
    KL coefficient $\beta$ & 0.001   & 0.001 \\
    \bottomrule
    \end{tabular}%
  \label{tab:hyperparameter}%
\end{table*}%

\section{Reproduction List}
In this section, we present a brief reproduction list to implement our method:
\begin{enumerate}
    \item \textbf{Computational Cards: } We use 4xA100 GPUs to train the Qwen3-4B-Base and Llama3.1-8B-Instruct for 4 days. We use 4xA100 GPUs to train the OctoThinker-8B-Hybrid-Base for 7 days. We use 8xA100 GPUs to train the Qwen3-8B-Base for 3 days.
    \item \textbf{Code: } We attach the implementation code in the supplementary materials.
    \item \textbf{Data: } All the dataset is officially available through their released links.
\end{enumerate}
\section{Related Work}
\label{sec:related_work_app}

This appendix complements the preliminary discussion in \S\ref{sec:prelim} by positioning our study in the broader landscape of  (i) data-efficient policy optimization under sparse/binary outcome rewards, (ii) prior attempts to resolve the diminishing-advantage phenomenon induced by homogeneous rollout groups, and (iii) intrinsic-reward designs based on model uncertainty.

\paragraph{Critic-Free Policy Optimization and Group-Based Advantages}
Policy-gradient methods~\citep{williams1992simple} constitute the foundation of RL-based fine-tuning, while PPO~\citep{schulman2017proximal} stabilizes updates via clipped importance ratios and KL regularization~\citep{kullback1951information}.
GRPO~\citep{shao2024deepseekmath} adapts PPO-style updates to a group sampling scheme by estimating advantages from the relative reward statistics within a set of rollouts generated from the same prompt.
This design eliminates a learned value critic and is thus memory-friendly for large-scale LLM training.
However, when all rollouts in a group share the same verifiable reward (all-correct or all-wrong), group-relative normalization yields vanishing advantages, directly linking optimization progress to the diversity of outcome rewards.

\paragraph{Data Efficiency in RLVR and Rollout Waste}
A central challenge in RLVR is data efficiency: each prompt requires sampling multiple rollouts to construct a useful advantage baseline, and sparse outcome rewards can cause many rollouts to contribute negligible learning signals.
As LLM capabilities improve~\citep{kaplan2020scaling,xiao2025densing}, an increasing fraction of prompts becomes trivially solvable, making \emph{positive homogeneous} (PH) groups more frequent and amplifying useless rollout costs.
DAPO~\citep{yu2025DAPO} targets this inefficiency through over-sampling prompts and filtering zero-advantage groups, which can increase the effective gradient density but still spends compute on rollouts that are later discarded.
Our work instead aims to make the sampled rollouts themselves more informative under PH groups, thereby improving utilization without adding extra rollouts.

\paragraph{Mitigating Negative Homogeneous Prompts (NH)}
A rich literature has explored how to reduce prompt difficulty or reshape the training signal for \emph{negative homogeneous} (NH) groups, where all sampled rollouts are incorrect.
Representative strategies include appending hints to prompts~\citep{liu2025ghpo}, adding in-context demonstrations~\citep{bamba2025xrpo}, and using replay buffers or replay-style mechanisms~\citep{sun2025improving,jiang2025vcrl}.
These approaches can convert homogeneous failures into heterogeneous outcomes by making at least some rollouts correct, yielding non-zero advantages.
We view NH-oriented techniques as largely complementary to our focus: our method targets the increasingly dominant PH regime and can in principle be combined with NH mitigation when needed.

\paragraph{Leveraging Positive Homogeneous Prompts (PH)}
In contrast to NH, dedicated treatments for PH groups remain comparatively under-explored, despite their increasing prevalence in modern RLVR pipelines.
Existing approaches typically fall into two categories.
(1) \textbf{Sampling-based heuristics.} DAPO~\citep{yu2025DAPO} filters out zero-advantage groups after over-sampling, improving effective batch quality at the expense of wasted rollouts.
(2) \textbf{Denser supervision via learned reward signals.} Implicit process reward models (PRMs)~\citep{yuan2025free,fei2025self} attempt to provide step-wise or token-level guidance beyond the binary outcome reward, but often rely on an SFT-tuned model or additional training/maintenance costs, which can inhibit adoption in strict zero-RL or large-scale settings.
Hybrid frameworks that cooperate with strong reward models~\citep{tao2025hybrid} can further enhance feedback richness, but likewise introduce extra compute and system complexity.
Our work targets PH groups while avoiding additional rollouts and dependence on large learned reward models.

\paragraph{Uncertainty Signals in RLVR}
Several recent RLVR studies incorporate model uncertainty as a lightweight intrinsic signal to enrich learning under sparse (often binary) verifiable rewards: (i) uncertainty-aware advantage shaping methods~\citep{xie2025unlocking} modulate GRPO-style updates using confidence/uncertainty at the response and token levels to improve exploration and credit assignment; (ii) intrinsic confidence-driven variants~\citep{wang2025icpo} turn relative confidence among multiple rollouts into a group-relative preference/advantage signal to augment the outcome reward signal; and (iii) calibration-oriented RLVR extensions~\citep{damani2025beyond,stangel2025rewarding} augment correctness with proper scoring-rule–based rewards so the model learns to output calibrated confidence alongside answers.
However, the first two paradigms overlook the advantage shaping of PH trajectories and fail to utilize them, while calibrated methods destroy the objective of the maximization of correctness, and achieve bad performance compared with pure RLVR baselines.
Our algorithm, which calibrates only on PH prompts, operates in a basically orthogonal direction with these methods, and would result in a further superior RLVR method when complementing these algorithms.


\begin{table*}[tbp]
  \centering

         \resizebox{\linewidth}{!}{%
    \begin{tabular}{lcccccccccccccc}
    \toprule
    \multirow{2}[4]{*}{\textbf{Model}} & \multicolumn{2}{c}{\textbf{AIME2024}} & \multicolumn{2}{c}{\textbf{AIME2025}} & \multicolumn{2}{c}{\textbf{AMC23}} & \multicolumn{2}{c}{\textbf{HMMT25}} & \multicolumn{2}{c}{\textbf{MATH}} & \multicolumn{2}{c}{\textbf{OlympiadB.}} & \multicolumn{2}{c}{\textbf{Avg.}} \\
\cmidrule(r){2-3} \cmidrule(r){4-5} \cmidrule(r){6-7} \cmidrule(r){8-9} \cmidrule(r){10-11} \cmidrule(r){12-13} \cmidrule(r){14-15}          & \textbf{P@1} & \textbf{P@K} & \textbf{P@1} & \textbf{P@K} & \textbf{P@1} & \textbf{P@K} & \textbf{P@1} & \textbf{P@K} & \textbf{P@1} & \textbf{P@K} & \textbf{P@1} & \textbf{P@K} & \textbf{P@1} & \textbf{P@K} \\
    \midrule
    GRPO  & 19.79 & 63.33 & 20.34 & 63.33 & 61.89 & \textbf{97.50} & 7.86  & 33.33 & 83.71 & 94.00 & 52.19 & 72.63 & 40.97 & 70.69 \\
    \rowcolor[rgb]{ .867,  .922,  .969} \our  & \textbf{25.86} & \textbf{73.33} & \textbf{22.97} & 60.00 & \textbf{69.65} & \textbf{97.50} & \textbf{10.81} & \textbf{46.67} & \textbf{86.93} & {95.80} & \textbf{57.07} & 76.25 & \textbf{45.55} & \textbf{74.92} \\
    \quad \textit{w/o} IR$^{\dagger}$ & 15.49 & 53.33 & 12.19 & 33.33 & 57.36 & 90.00 & 2.71  & 20.00 & 79.64 & 91.80 & 46.59 & 68.50 & 35.66 & 59.49 \\
    \quad \textit{w/o} FW  & 24.43 & 66.67 & 21.09 & \textbf{66.67} & {68.81} & 95.00 & 10.39 & 36.67 & 85.93 & 95.20 & 55.26 & \textbf{76.76} & 44.32 & 72.83  \\
    \quad \textit{w/o} AC$^{\dagger}$ & 20.23 & 66.67 & 19.48 & 53.33 & 60.55 & \textbf{97.50} & 7.89  & 33.33 & 82.95 & \textbf{96.00} & 50.87 & 73.67 & 40.33 & 70.08 \\
    \bottomrule

    \end{tabular}%
    }
      \caption{Ablation study on \our. We compare (1) without intrinsic reward~(w/o IR); (2) without focal weight~(w/o FW); and (3) without advantage calibration~(w/o AC) to unveil that each design of \our are beneficial for simultaneously enhanced Pass@1 and Pass@K scores. The experiments marked with a $\dagger$ failed to complete training; we used the checkpoint saved before the crash for testing.}
  \label{tab:ablation}%
\end{table*}%

\section{Experimental Details}

\subsection{Descriptions of Math Testbeds}
\label{sec:desc_testbeds}
We present the detailed description of the mathematical evaluation datasets as follows:
\begin{enumerate}[leftmargin=*]
    \item \textbf{AIME2024, AIME2025}~\citep{aime2024,aime2025}: These two datasets contain High school Olympiad-level assessment from American Invitational Mathematics Examination in 2024 and 2025. Each dataset contains 30 challenging problems covering Algebra/Geometry/Number theory.
    \item \textbf{AMC23}~\citep{zwhe99_amc23}: This dataset is sourced from American Mathematics Competitions system in 2023, which contains 40 problems with hybrid question types.
    \item \textbf{OlympiadBench}~\citep{he-etal-2024-olympiadbench}: This dataset contains comprehensive math Olympiad problems from various nations. We only select the English version related to Math and keep the problems that require an answer with a number, leaving 581 problems for evaluation in total.
    \item \textbf{MATH500}~\citep{lightman2023let}: This dataset is an advanced mathematics evaluation set curated by OpenAI containing 500 problems with formal mathematical notations.
    \item \textbf{HMMT25}~\citep{balunovic_srimatharena_2025}: The original questions were sourced from the HMMT February 2025 competition. 30 questions were extracted, converted to LaTeX and verified.
\end{enumerate}

\subsection{Descriptions of Medical Testbeds}
\label{sec:desc_med_testbeds}
We present the detailed description of the medical evaluation datasets as follows:
\begin{enumerate}[leftmargin=*]
    \item \textbf{MedQA}~\citep{jin2021disease} is a widely used benchmark for evaluating AI systems in medical question answering, featuring multiple-choice questions from professional medical licensing exams such as the USMLE and exams from China and Taiwan. We adopt its 5-options English version, taking the 1,273 test problems as the evaluation benchmark.
    \item \textbf{PubmedQA}~\citep{jin-etal-2019-pubmedqa} is a specialized benchmark for biomedical question answering, consisting of question-answer pairs derived from PubMed abstracts. It focuses on yes/no/maybe questions that require reasoning over biomedical literature. We use the human-labeled question test set, with 500 problems for evaluation. Note that we include relevant contexts before questions, challenging models' reasoning capability among contexts.
    \item \textbf{MedMCQA}~\citep{pal2022medmcqa} is a large-scale benchmark for medical question answering, featuring over 194,000 multiple-choice questions sourced from Indian medical entrance exams and other educational resources. It spans a wide range of medical topics, including anatomy, pharmacology, and pathology, and is designed to evaluate the reasoning and knowledge application skills of AI systems in a clinical context. The test set contains 4,183 problems.
    \item \textbf{MMLU-Pro}~\citep{wang2024mmlupro} is a challenging multi-task benchmark containing over 12,000 multiple-choice questions across 14 diverse domains, including subjects in STEM (e.g., math, physics, chemistry), social sciences, law, and humanities. We only maintain \texttt{health} and \texttt{biology} subsets for testing medical reasoning abilities, which includes 1535 problems.
    \item \textbf{MedXpertQA}~\citep{zuo2025medxpertqa} is an expert-level medical benchmark comprising 4,460 questions spanning 17 medical specialties and 11 body systems. It includes two subsets: a text-only version for evaluating textual medical reasoning and a multimodal version (MM) with images, aimed at assessing advanced clinical knowledge comparable to medical licensing exams. We only test models on the text-only subset, which contains 2450 problems.
\end{enumerate}

\begin{figure*}[tbp]
    \begin{promptbox}{Training and evaluation prompt}
A conversation between User and Assistant. The user asks a question, and the Assistant solves it. The assistant first thinks about the reasoning process in the mind and then provides the user with the answer. The assistant thinks deeply and output the final answer within \textbackslash\textbackslash boxed\{\}. 

User: \{prompt\}

Assistant: 
\end{promptbox}
\caption{Training and evaluation prompt}
\label{fig:reason_prompt}

\end{figure*}

\subsection{Evaluation Prompts}
\label{sec:evaluation_prompt}
For the mathematical reasoning tasks, we use prompts defined in Fig.~\ref{fig:reason_prompt} to start reasoning.
For the medical reasoning tasks, We prompt the LRM with ``\texttt{Please reason step by step and output the final answer as `The answer is'} '' and extract the contents after `The answer is' to \textbf{exact-match} the ground truth answer.

\subsection{Computation of Metrics}
\label{sec:computation_metrics}

\paragraph{Pass@K} The pass@K~\citep{chen2021evaluating} scores are computed as below:
\begin{equation}
    \mathrm{pass@K}=1 - \frac{ \binom{n - c}{K} }{ \binom{n}{K} }
\end{equation}
where $n$ is the number of samples and $c$ is the number of correct samples.
When $K$ is set to 1, this metric is reduced to the average accuracy among the $n$ samples.

\subsection{Details of Ablation Study}
In this section, we in depth introduce the implementation of three variants of \our presented in \S\ref{sec:ablation}.
For \texttt{w/o IR}, we reward all the responses from homogeneous groups with a fixed advantage score (0.05). 
This value is taken by referring to \citet{zhu2025the}, which rewards positive rollouts with a value less than $0.1$.
For \texttt{w/o FW}, the weight $w_{i,j}$ for a token $o_{i,j}$ from a rollout $\vo_i$ is set to 1 for any token.
For \texttt{w/o AC}, the $\mathcal{A}_{i,j}^{\rm final}$ equals to the original $\tilde{A}_{i,j}$ without calibration.

\subsection{Details of Experiments of RQ2}
\label{sec:details_of_other_exploration}
These experiments involve many other baselines to enhance the model exploration.
The following baselines are fetched directly from their officially released codebase.
For BAPO, it controls the clip range in an asymmetric manner, which allows the clip\_high argument to be adjusted within the range $[1.5, 3.0]$ in a step of 0.1, and the clip\_low argument to be adjusted within $[0.5,0.95]$ in a step of 0.05.
The termination rule for adjustment is that the ratio of positive tokens accounts for 50\% of the training batch.
For KL-Cov, tokens with the top 0.2\% covariance would be augmented with a KL loss by regulating the distribution between the old policy and the current policy.
For Clip-Cov, the original clip range $[0.8, 1.2]$ is modified to $[0,2]$. Meanwhile, 0.02\% tokens whose covariance score is located within $[1.0,5.0]$ would be sampled randomly from the training batch and clipped from the current training step.
For Entropy-Adv, the additional entropy bonus, defined as $\psi(\mathcal{H}_{i,j})=\alpha\cdot\mathcal{H}_{i,j}$ ($\alpha=0.4$) is set to be no greater than half of the absolute value of the original advantage score. The shaped advantage function is defined as $A_{i,j}^{\rm shaped}=A_{i,j}+\psi(\mathcal{H}_{i,j})$.

\subsection{Details of Experiments of RQ5}
\label{sec:details_of_med_corpus}
The training data is constructed as follows.
As the original training set of MedMCQA contains 182K data, which includes many low-quality questions.
Therefore, we use a simple filtering rule, which we prompt Llama3.1-8B-Instruct~\citep{DBLP:journals/corr/abs-2407-21783} to conduct \textbf{greedy} decoding on the whole dataset, and filter questions that are judged to be correct.
We do not conduct a similar filtering process on the MedQA training set, as its data is more challenging than that of MedMCQA.
Finally, the number of training set from MedMCQA reaches 73K; the combination of the MedQA training set includes 85K high-quality training data.

\begin{figure*}[t]
    \centering
    \begin{subfigure}{0.32\linewidth}
        \centering
        \includegraphics[width=\linewidth]{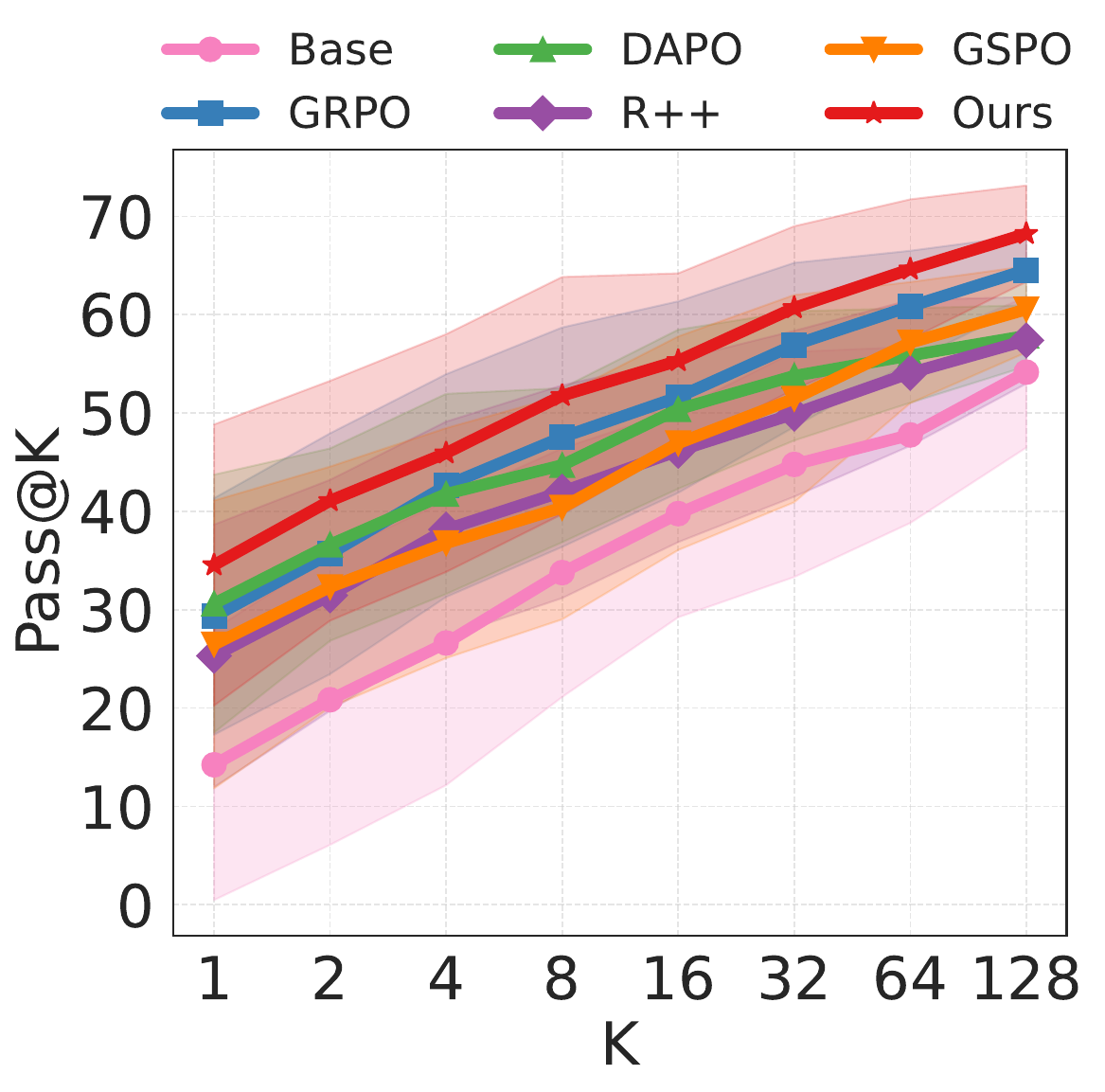}
        \caption{}
        \label{fig:passk_algorithm_cmp}
    \end{subfigure}
    \begin{subfigure}{0.32\linewidth}
        \centering
        \includegraphics[width=\linewidth]{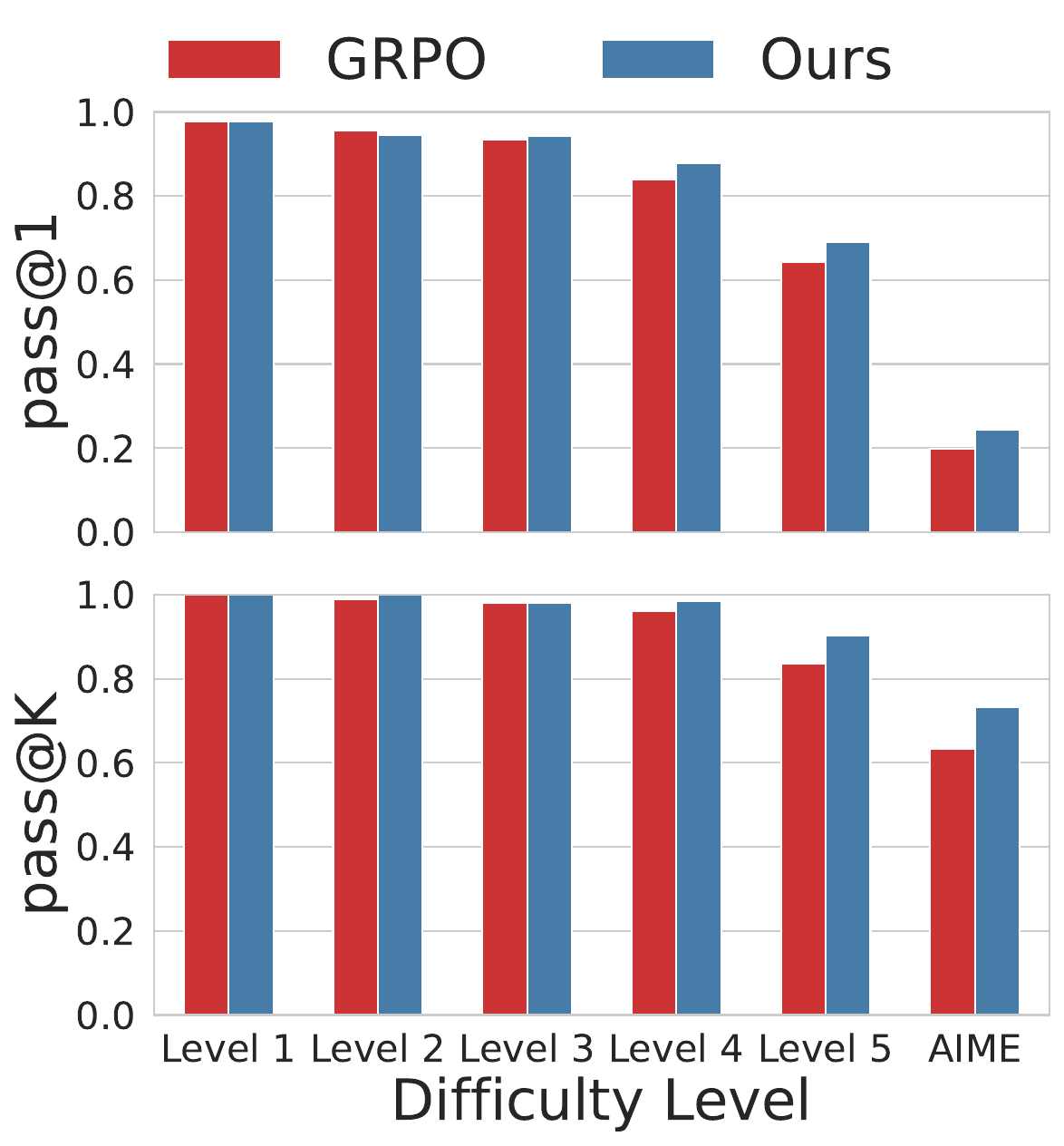}
        \caption{}
        \label{fig:difficulty}
    \end{subfigure}%
    \begin{subfigure}{0.32\linewidth}
        \centering
        \includegraphics[width=\linewidth]{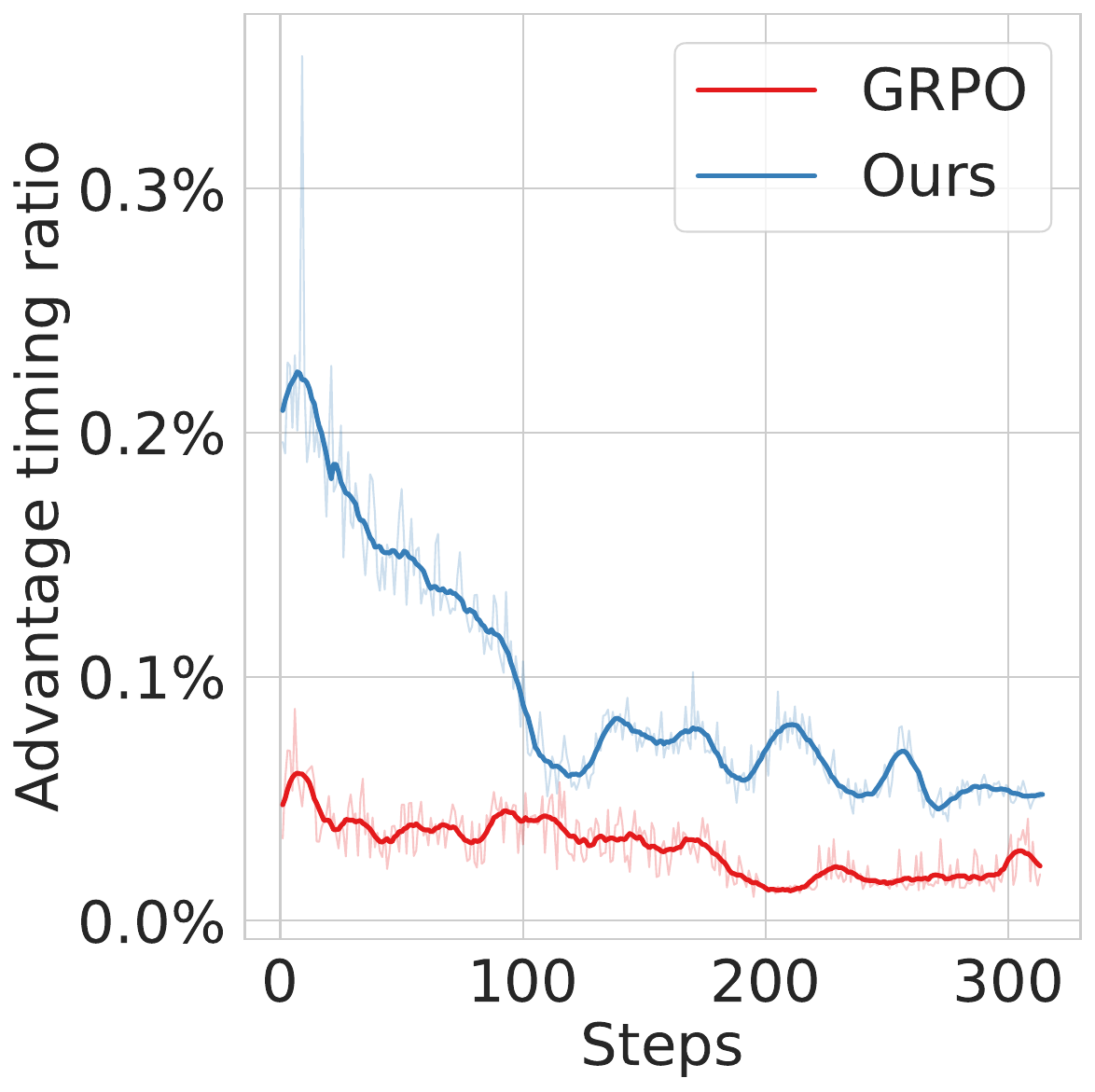}
        \caption{}
        \label{fig:timing_cmp}
    \end{subfigure}
    \caption{(a) Pass@K scaling comparison of \our against other GRPO variants; (b) Maintain performance on easy queries and breaking boundaries on challenging problems when evaluating \our on six difficulty levels sourced from MATH500 and AIME2024; (c) Negligible extra computational overhead compared to GRPO.}
    \label{fig: }

\end{figure*}

\section{Additional Experiments}

\subsection{Pass@K Scaling}
We compare \our with the base model and GRPO variants, on the representative exploration metric, i.e., Pass@K, to unveil the improved exploration potentials.
Specifically, we select $k=[1,2,4,8,16,32,64,128]$, and show the detailed improvements on the four challenging benchmarks (AIME24, AIME25, AMC23, and HMMT25) in Table-[\ref{tab:aime24_scale_table},\ref{tab:aime25_scale_table},\ref{tab:amc23_scale_table},\ref{tab:hmmt25_scale_table}] with 10 runs.
The visualized result is shown in Fig.~\ref{fig:passk_algorithm_cmp}.
We observe that \our consistently outperforms other methods in most of the challenging benchmarks under various sampling candidates, especially on the extremely challenging HMMT25 set, which reflects the superb potential of \our for breaking the capability boundary.

\subsection{Performance Across Diverse Difficulties}
In this section, we examine whether \our improves the model comprehensively across diverse difficulty levels.
We compute pass@1 and pass@K scores on the MATH500 and AIME2024 datasets, which provide self-contained difficulty gradients across six levels.
As shown in Fig.~\ref{fig:difficulty}, \our maintains the same mastery as GRPO on easy problems, again verifying that \our does not sacrifice exploitation for exploration.
And the performance leap enlarges with the increase of problem complexity, showing \our enhances LRM's performance in a promising manner for breaking more knowledge boundaries.

\subsection{Computational Overhead}
In this section, we compare \our with GRPO in terms of the timing cost of the auxiliary advantage computation.
We derive the ratio of time for computing advantages against that for completing a training step during a whole training epoch.
Results in Fig.~\ref{fig:timing_cmp} demonstrate that the additional advantage incurs less than 0.05\% more timing cost than the normal GRPO baseline.
This consolidates our claim that \our adds near-zero computation while being sufficiently effective across models and tasks.

\begin{table*}[tbp]
  \centering

           \resizebox{\linewidth}{!}{%
    \begin{tabular}{lcccccccccccccc}
    \toprule
    \multirow{2}[4]{*}{$\mathbf{\lambda_{\max}}$} & \multicolumn{2}{c}{\textbf{AIME2024}} & \multicolumn{2}{c}{\textbf{AIME2025}} & \multicolumn{2}{c}{\textbf{AMC23}} & \multicolumn{2}{c}{\textbf{HMMT25}} & \multicolumn{2}{c}{\textbf{MATH}} & \multicolumn{2}{c}{\textbf{OlympiadB.}} & \multicolumn{2}{c}{\textbf{Avg.}} \\
\cmidrule{2-15}          & \textbf{P@1} & \textbf{P@K} & \textbf{P@1} & \textbf{P@K} & \textbf{P@1} & \textbf{P@K} & \textbf{P@1} & \textbf{P@K} & \textbf{P@1} & \textbf{P@K} & \textbf{P@1} & \textbf{P@K} & \textbf{P@1} & \textbf{P@K} \\
    \midrule
    0 (GRPO) & 19.79 & 63.33 & 20.34 & \textbf{63.33} & 61.89 & 97.50 & 7.86  & 33.33 & 83.71 & 94.00 & 52.19 & 72.63 & 40.96 & 70.69 \\
    $1e-3$ & 21.74 & 63.33 & 19.22 & 56.67 & 65.98 & 95.00 & 9.74  & 33.33 & 84.89 & 94.80 & 53.03 & 73.32 & 42.43 & 69.41 \\
    \rowcolor[rgb]{ .867,  .922,  .969} $1.5e-3$ & \textbf{25.86} & \textbf{73.33} & \textbf{22.97} & 60.00 & \textbf{69.65} & \textbf{97.50} & \textbf{10.81} & \textbf{46.67} & \textbf{86.93} & {95.80} & \textbf{57.07} & \textbf{76.25} & \textbf{45.55} & \textbf{74.92} \\
    $2e-3$ & 20.21 & 66.67 & 19.19 & 60.00 & 61.88 & 97.50 & 10.13 & 43.33 & 83.80 & \textbf{96.60} & 52.02 & 75.04 & 41.20 & 73.19 \\
    \bottomrule
    \end{tabular}%
    }
      \caption{Hyperparameter sensitivity analysis on $\lambda_{\max}$. $1e-3$ renders slow convergence given limited data, while $2d-3$ would interfere but not benefit the objective of maximizing correctness. }
  \label{tab:sensitivity}%
\end{table*}%
\subsection{Sensitivity Analysis}
\label{sec:sensitivity_analysis}
In this section, we study the sensitivity of the only hyperparameter $\lambda_{max}$ using by \our.
We use grid-search using three different $\lambda_{\max}$ values: [1e-3, 1.5e-3, 2e-3] on Qwen3-4B-Base model using the same training configurations used in \S\ref{sec:exp_setup}.
The results in Table~\ref{tab:sensitivity} demonstrate that $1e-3$ is stable but results in a slow convergent rate, while the auxiliary learning signal given $\lambda_{\max}=2e-3$ pathologically impacts the major optimization objective.
In contrast, the value $1.5e-3$ is modest and suitable for both a stable training procedure and fast convergence, presenting significant improvements on Pass@1 and Pass@K with only $\sim$300 RL updates.
Meanwhile, this hyperparameter is extendable to a larger model, Qwen3-8B-Base and even Llama with a different backbone and intelligence, demonstrating that our method is not sensitive.

\begin{figure*}[tbp]
    \centering
    \includegraphics[width=\linewidth]{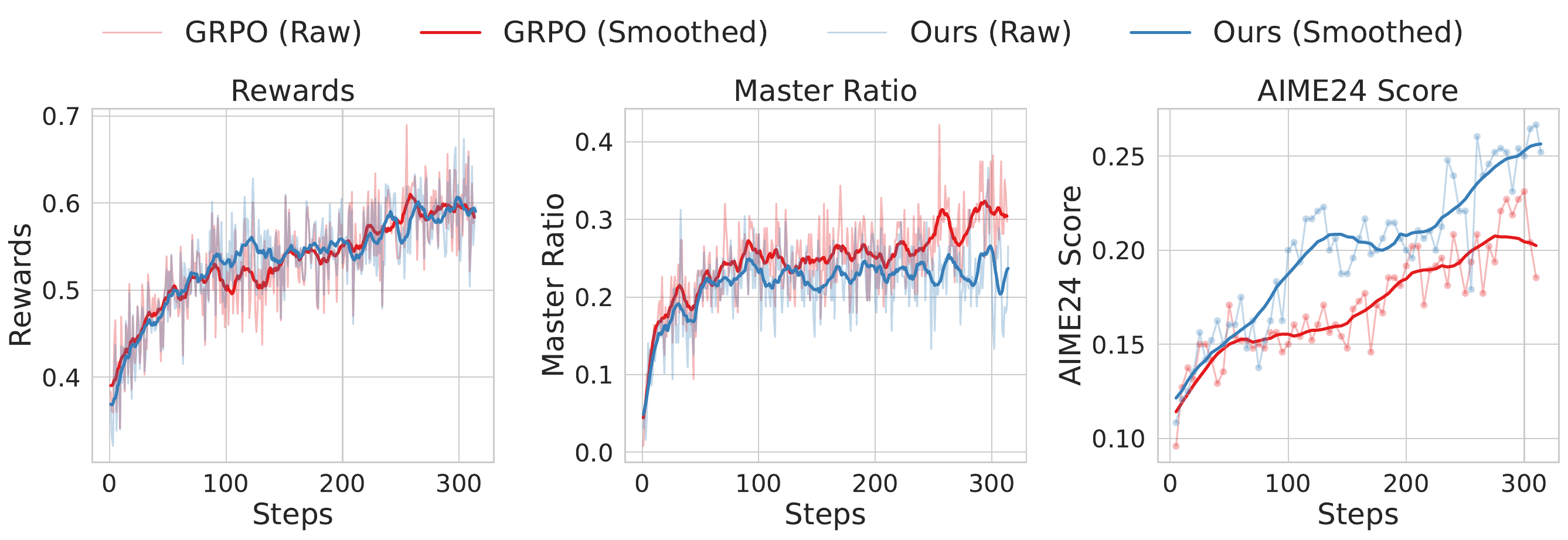}
    \caption{Training rewards, master ratio (PH ratio) and AIME24 dev set score of GRPO and our method trained with Qwen3-4B-Base on the mathematical reasoning task.}

    \label{fig:qwen34b_log}

\end{figure*}
\begin{figure*}[tbp]
    \centering
    \includegraphics[width=\linewidth]{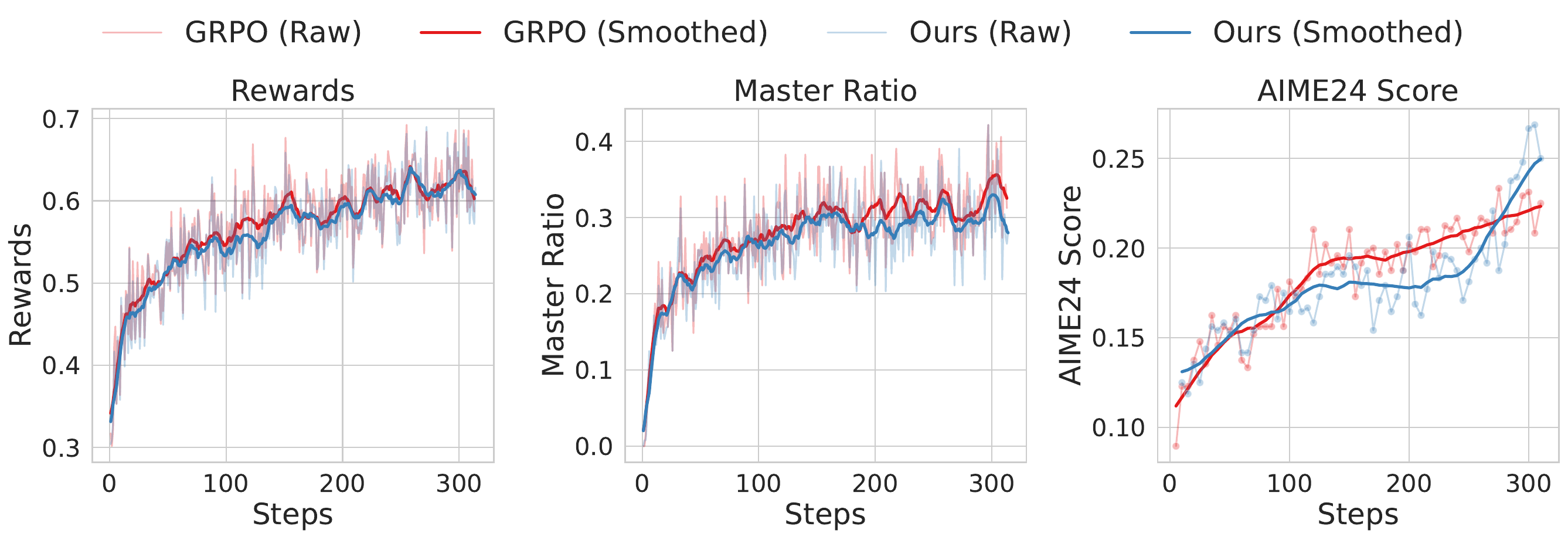}
    \caption{Training rewards, master ratio (PH ratio) and AIME24 dev set score of GRPO and our method trained with Qwen3-8B-Base on the mathematical reasoning task.}

    \label{fig:qwen38b_log}

\end{figure*}
\begin{figure*}[tbp]
    \centering
    \includegraphics[width=\linewidth]{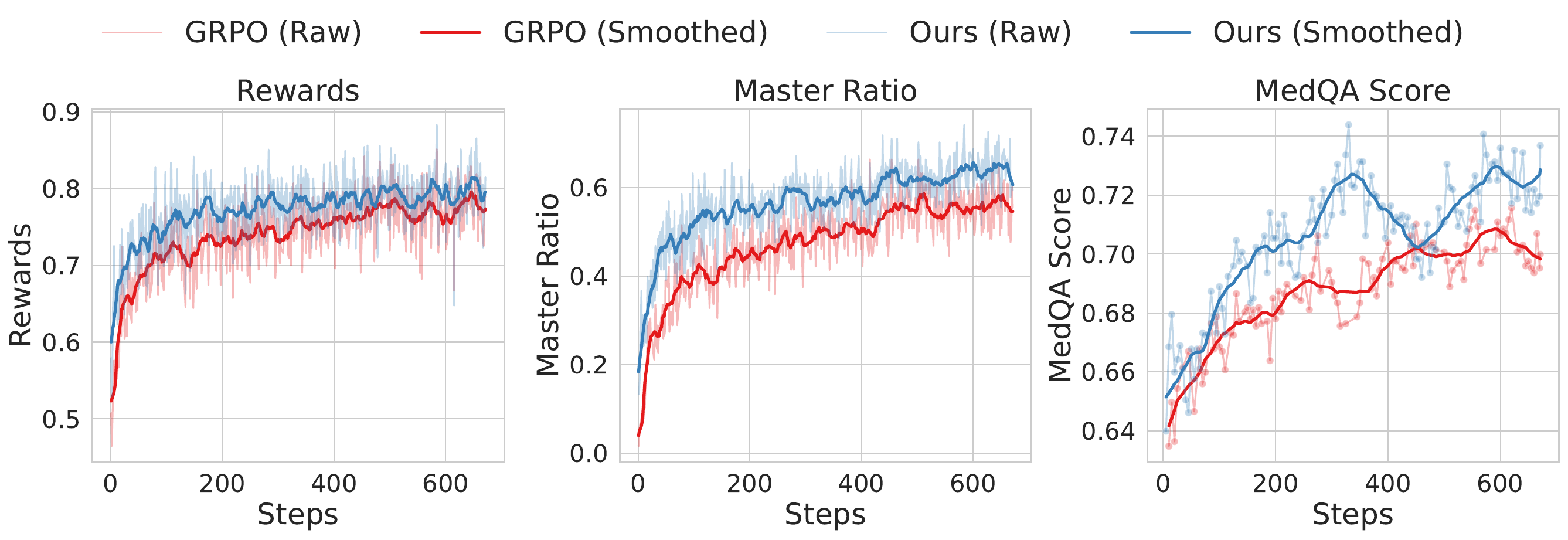}
    \caption{Training rewards, master ratio (PH ratio) and MedQA dev set score of GRPO and our method trained with Llama3.1-8B-Instruct on the medical reasoning task.}

    \label{fig:med_log}

\end{figure*}

\subsection{Training Logs}
We present the training logs of mathematical reasoning, by taking Qwen3-4B-Base and Qwen3-8B-Base in Fig.~\ref{fig:qwen34b_log} and Fig.~\ref{fig:qwen38b_log}, respectively.
For the medical reasoning, we present training logs with Llama3.1-8B-Instruct as the backbone in Fig.~\ref{fig:med_log}.
Here, ``master ratio'' is the ratio of PH prompts within a batch.

\begin{table*}[tbp]
  \centering
       \resizebox{\linewidth}{!}{%
    \begin{tabular}{lcccccccccccccc}
    \toprule
    \multirow{2}[4]{*}{\textbf{Model}} & \multicolumn{2}{c}{\textbf{AIME2024}} & \multicolumn{2}{c}{\textbf{AIME2025}} & \multicolumn{2}{c}{\textbf{AMC23}} & \multicolumn{2}{c}{\textbf{HMMT25}} & \multicolumn{2}{c}{\textbf{MATH}} & \multicolumn{2}{c}{\textbf{OlympiadB.}} & \multicolumn{2}{c}{\textbf{Avg.}} \\
\cmidrule(r){2-3} \cmidrule(r){4-5} \cmidrule(r){6-7} \cmidrule(r){8-9} \cmidrule(r){10-11} \cmidrule(r){12-13} \cmidrule(r){14-15}          & \textbf{P@1} & \textbf{P@K} & \textbf{P@1} & \textbf{P@K} & \textbf{P@1} & \textbf{P@K} & \textbf{P@1} & \textbf{P@K} & \textbf{P@1} & \textbf{P@K} & \textbf{P@1} & \textbf{P@K} & \textbf{P@1} & \textbf{P@K} \\
    \midrule

    \multicolumn{15}{c}{\textit{Base Model: OctoThinker-8B-Hybrid-Base (Llama Arch.)}} \\
    \midrule
    Base  & 0.96  & 26.67 & 0.49  & 23.33 & 18.95 & 92.50 & 0.18  & 13.33 & 32.63 & 82.00 & 18.18 & 51.29 & 11.90 & 48.19 \\
    GRPO  & 11.95 & 43.33 & 12.01 & \textbf{50.00} & 47.70 & 92.50 & 4.64  & \textbf{40.00} & 76.06 & 93.00 & 42.96 & 67.47 & 32.55 & 64.38 \\
    DAPO  & 12.03 & 46.67 & 10.05 & 43.33 & 52.05 & 92.50 & 4.90  & 20.00 & 75.93 & 91.80 & 43.58 & 65.92 & 33.09 & 60.04 \\
    R++   & 12.53 & 36.67 & 9.32  & 46.67 & 48.42 & 92.50 & 3.49  & 26.67 & 75.98 & \textbf{93.40} & 41.91 & 67.64 & 31.94 & 60.59 \\
    GSPO  & 8.36  & 43.33 & 7.50  & 36.67 & 41.43 & 85.00 & 3.91  & 23.33 & 66.51 & 87.40 & 33.87 & 58.52 & 26.93 & 55.71 \\
    \rowcolor[rgb]{ .867,  .922,  .969} \our  & \textbf{14.11} & \textbf{50.00} & \textbf{13.28} & \textbf{50.00} & \textbf{52.73} & \textbf{97.50} & \textbf{5.26} & 36.67 & \textbf{77.16} & \textbf{93.40} & \textbf{45.80} & \textbf{69.19} & \textbf{34.73} & \textbf{66.13} \\
    \bottomrule

    \end{tabular}%
    }
      \caption{Comprehensive comparison against other critic-free RL algorithms in terms of Pass@1 (P@1) and Pass@K (P@K) scores. ``OlympiadB.'' refers to the OlympiadBench. Best performance is highlighted with \textbf{bold}.}
  \label{tab:octo_table}%
\end{table*}%

\subsection{Trial of RL on Math with Llama}
\label{sec:trial_on_llama}
It is difficult to apply \our to Llama, for the following two reasons: (i) Due to lack of necessary pre-training data, Llama models fail to incentivize reasoning abilities as Qwen; (ii) Even if Llama models are injected required corpus via mid-training~\citep{mo2025mid}, its reasoning ability is much lower than models using the Qwen backbone, resulting in a much lower portion of PH prompts among the batch.
The extreme case where the portion of PH prompts decreases to 0 degrades \our to normal GRPO.
These two reasons lead to not as significant improvement gains as applying \our to Qwen architectures when optimizing math reasoning.
To illustrate, we choose OctoThinker-8B-Hybrid-Base~\citep{wang2025octothinker}, which undergoes a fine-grained mid-training procedure to enable incentivization of necessary reasoning capabilities like Qwen.
Results in Table~\ref{tab:octo_table} demonstrate that \our still outperforms the other algorithms, but with a relatively small margin as expected.
However, \our still makes \textbf{+2.18} gains on Pass@1 and \textbf{+1.75} gains on Pass@K compared to GRPO, and even outperform the DAPO baseline, with 1.64 absolute gains in Pass@1 and 6.09 pass@K gains in Pass@K, whose rollouts receive non-zero advantages.
Given that \our's strong performance in such a disadvantageous scenario, it is sufficiently generalizable to other modern models with even higher intelligence.

\begin{figure*}[t]
    \centering
    \includegraphics[width=\linewidth]{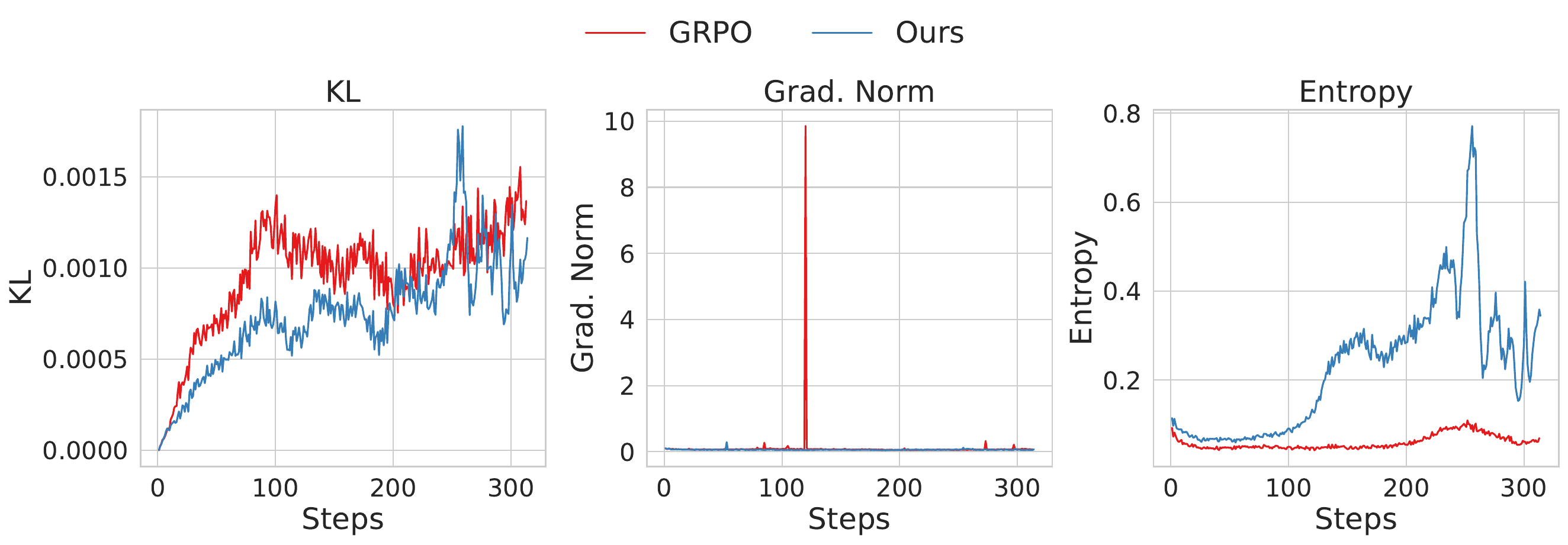}
    \caption{The KL loss, gradient norm and entroy dynamics of applying \our and GRPO algorithms built on  Qwen3-4B. After undergoing a long range of exploration, with steadily increasing policy entropy values, our method quickly transforms the exploration to exploitation signals, accompanied by a rapid fall back of entropy signals and a surge in performances of downstream benchmarks (see Fig.~\ref{fig:qwen34b_log}).}

    \label{fig:qwen34b_stability}

\end{figure*}
\begin{figure*}[t]
    \centering
    \includegraphics[width=\linewidth]{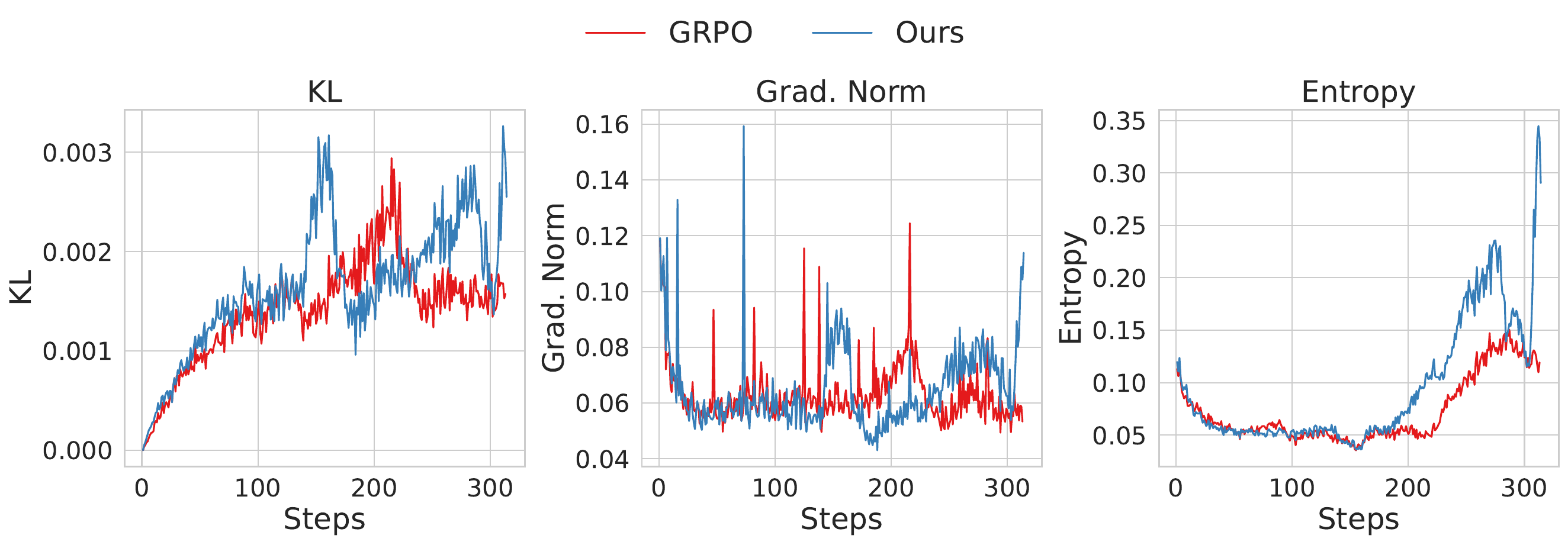}
    \caption{The KL loss, gradient norm and entroy dynamics of applying \our and GRPO algorithms built on  Qwen3-8B. After undergoing a long range of exploration, with steadily increasing policy entropy values, our method quickly transforms the exploration to exploitation signals, accompanied by a rapid fall back of entropy signals and a surge in performances of downstream benchmarks (see Fig.~\ref{fig:qwen38b_log}).}

    \label{fig:qwen38b_stability}

\end{figure*}

\subsection{Training Stability}
Due to severe computational resource constraints, all methods including our approach and the other baselines were trained for a fixed duration of 314 steps (equivalent to one full epoch of data collection). We acknowledge that this may raise concerns about whether the algorithms have fully converged. However, three key observations support the robustness of our conclusions:

First, the training dynamics of our method exhibit exceptional stability. As shown in Fig.~[\ref{fig:qwen34b_stability},\ref{fig:qwen38b_stability}], the KL divergence remained consistently low, and the gradient norm stayed within an even narrower range, indicating no signs of divergence or oscillation. This stability implies that extending training would likely preserve our method’s performance gains rather than erode them.

Second, the late-stage performance surge (observed in the final 30 steps) is not an artifact of under-training but reflects our method’s deliberate exploration-exploitation trade-off. Specifically, the algorithm prioritizes extensive exploration of the policy space in early stages (evidenced by steady entropy enhancement), enabling it to discover high-reward regions that GRPO overlooks. Once a promising mode is identified (around step 280), rapid policy refinement occurs, causing the sharp performance lift. This behavior, common in entropy-regularized RL algorithms, is a feature of \our; it ensures thorough exploration before committing to exploitation.
This behavior helps to achieve a great trade-off between pass@1 and pass@K, avoiding mode collapse which is a known drawback of GRPO.

Consequently, while longer training was infeasible for all baselines under our constraints, the combination of stable convergence indicators and the intrinsic exploration dynamics suggests our method would maintain its lead if trained to full convergence. We will extend the training to guarantee a sound conclusion in the later stage.

\begin{figure*}[tbp]
    \centering
    \includegraphics[width=\linewidth]{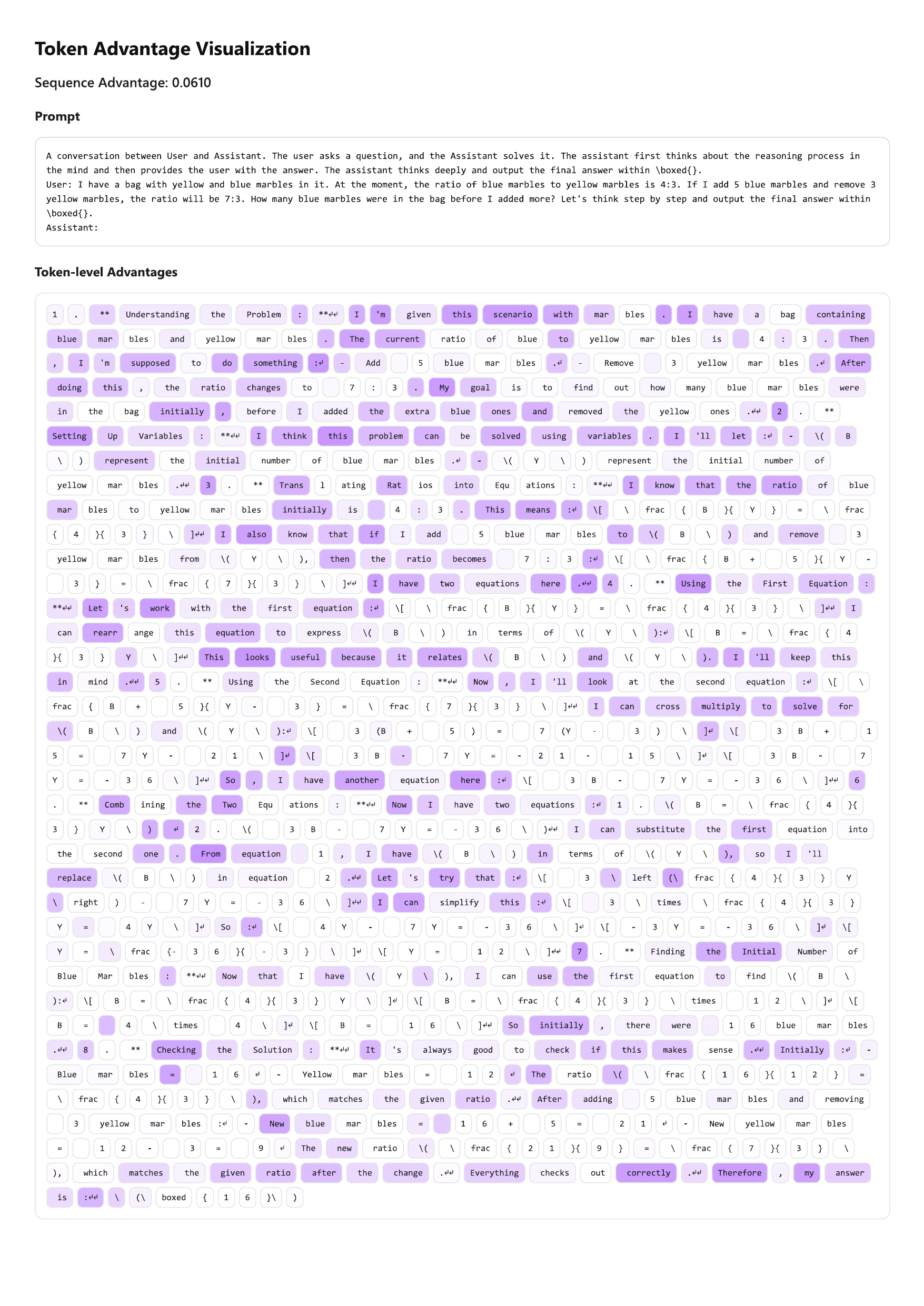}
    \caption{Advantage distribution of \our. The problem is sourced from the MATH500 dataset. The darker the token, the more advantage credit is assigned. The maximum token advantage equals to the sequence advantage value.}

    \label{fig:case}

\end{figure*}




\section{Clarification of \our on Trivial Prompts}
In this section, we clarify that MINER naturally skips \textit{truly trivial} problems (e.g., 1+1) while selectively reinforcing \textit{fragile-correct} PH prompts where correctness is unstable—a distinction critical to data efficiency.
\begin{enumerate}
    \item \textbf{Automatic filtering of trivial problems.} By omitting std-normalization in Eq.~\ref{eq:miner_adv_init} (unlike GRPO Eq.~\ref{grpo_adv}), \our preserves absolute uncertainty scales. For trivial prompts like 1+1, all rollouts exhibit near-identical low NLL, which implies that $A_i^{\rm int}\approx 0$ after mean-centering and the rectified value $\mathrm{ReLU}(A_i^{\rm int})\approx 0$. Token-level focal weighting~(Eq.~\ref{eq:focal_weight}) further suppresses gradients on deterministic tokens. Thus, \our automatically allocates zero gradient budget to already-mastered prompts, requiring no manual filtering.
    \item \textbf{Value of ``fragile-correct'' PH prompts.} 100\% accuracy within 16 trials is unequal to true mastery. As shown in Fig.~\ref{fig:difficulty}, \our gains 0 improvements on Level 1--3 (truly mastered) but +5.2\% on Level 4--6 (fragile-correct) compared to GRPO, confirming it targets unstable correctness instead of already mastery. Crucially, training curves (Fig.~\ref{fig:qwen34b_log}-\ref{fig:qwen38b_log}) reveal a late-stage surge (>+5 Pass@1 in the final 30 steps), evidencing that consolidating fragile paths yields delayed but high-value signals instead of trivial memorization.
    \item \textbf{Generalization beyond training domain.} MINER's gains transfer to medical reasoning (Table~\ref{tab:medical_cmp}) and scale with test-time compute (Fig.~\ref{fig:maj_at_k}, \ref{fig: passk_scale}), demonstrating that PH consolidation enhances robust reasoning capability rather than overfitting. In Fig.~\ref{fig:med_log}, the policy answers 60\% of the problems with all 16 rollouts. This is similar to a scenario where input problems are easy enough for the policy. However, from Table~\ref{tab:medical_cmp}, \our consistently improves over GRPO by 1.39 pass@1 and 1.84 pass@K on average, which demonstrates that with the proposed three key components of Miner, it can learn only useful information from PH prompts and boost the final performance in terms of exploration and exploitation even trained on a large amount of easy problems.
    \item \textbf{Connection with prior work.} \our extracts additional signals from the policy's intrinsic uncertainty apart from the outcome verifier, without external synthesis or auxiliary models. Such intrinsic learning signals impart valuable learning signals and the policy could learn from positive homogeneous problems, i.e., problems whose all rollouts are incorrect.
\end{enumerate}


\section{License}
\paragraph{License} We plan to release our training and evaluation code under the MIT License (or Apache-2.0). Model checkpoints will be distributed for research use only and will comply with the license terms of the underlying base model.

\paragraph{Intended use and compatibility with upstream terms.} We use existing artifacts (e.g., base models, datasets, and toolchains) in a manner consistent with their stated intended use and license/terms when specified. In particular, we only use resources available for research and comply with any restrictions on redistribution and derivative works.
We will release our code under the [MIT/Apache-2.0] license. We will release our model checkpoints for research use only, and their use and redistribution are subject to the licenses/terms of the underlying base model(s) and dataset(s). We do not authorize uses that would violate upstream access conditions (e.g., non-research use when restricted) and we do not claim additional rights over third-party resources.

\section{Data Cleaning}
We use only publicly available open-source datasets. These datasets consist of task/problem content and are not intended to contain personally identifying information. We additionally performed sanity checks via automated pattern matching (e.g., emails/phone numbers/URLs) and spot-checking, and did not observe PII. We do not redistribute any third-party data and only release code/model under the upstream licenses/terms; any examples shown are sanitized.

\section{Case Study}
In this section, we present the advantage distribution of \our trained on Qwen3-8B-Base. 
We randomly choose one problem sourced from MATH500 to illustrate the advantage score, as it contains numerous PH prompts.
The case shown in Fig.~\ref{fig:case} demonstrates that most deterministic tokens are not rewarded, which prevents overfitting to already-mastered actions.
Noticeably, \our enhances the frequency of rewarding reasoning-required actions, including ``Checking'' for verification, ``So'' for implication, ``Let'' for solution progression, and ``Combining'' for conditions gathering.
This case reflects meaningful learning signals for utilizing PH prompts via our method, significantly enhancing the data efficiency for RLVR.

\begin{table*}[tbp]
\centering
\resizebox{\textwidth}{!}{%
\begin{tabular}{l *{8}{c}}
\toprule
\textbf{Model} & \textbf{1} & \textbf{2} & \textbf{4} & \textbf{8} & \textbf{16} & \textbf{32} & \textbf{64} & \textbf{128} \\
\midrule
Base         & $8.00 \pm 9.19$ & $12.67 \pm 8.92$ & $17.67 \pm 9.25$ & $22.33 \pm 10.57$ & $28.33 \pm 10.22$ & $36.67 \pm 12.92$ & $37.67 \pm 12.15$ & $45.00 \pm 8.22$ \\
GRPO         & $22.00 \pm 10.02$ & $24.67 \pm 9.72$ & $28.33 \pm 12.88$ & $31.00 \pm 9.97$ & $38.00 \pm 13.88$ & $43.33 \pm 15.22$ & $51.67 \pm 10.90$ & $57.00 \pm 7.55$ \\
DAPO         & $22.33 \pm 12.35$ & $24.33 \pm 6.19$ & $29.67 \pm 8.79$ & $36.33 \pm 10.57$ & $40.00 \pm 9.22$ & $45.33 \pm 9.65$ & $50.33 \pm 3.53$ & $52.33 \pm 1.53$ \\
REINFORCE++  & $15.00 \pm 7.53$ & $19.33 \pm 9.49$ & $24.33 \pm 9.30$ & $27.33 \pm 10.39$ & $31.67 \pm 13.18$ & $39.33 \pm 13.06$ & $46.00 \pm 12.04$ & $51.00 \pm 9.16$ \\
GSPO         & $15.67 \pm 9.25$ & $20.00 \pm 6.67$ & $25.00 \pm 10.63$ & $27.67 \pm 13.69$ & $32.67 \pm 13.63$ & $40.00 \pm 15.87$ & $51.33 \pm 18.48$ & $61.33 \pm 13.49$ \\
\rowcolor[rgb]{ .867,  .922,  .969} \our & $\mathbf{25.67} \pm 14.08$ & $\mathbf{30.33} \pm 11.39$ & $\mathbf{36.67} \pm 13.86$ & $\mathbf{43.33} \pm 17.08$ & $\mathbf{45.67} \pm 8.19$ & $\mathbf{54.67} \pm 10.02$ & $\mathbf{61.00} \pm 9.18$ & $\mathbf{67.00} \pm 5.97$ \\
\bottomrule

\end{tabular}
}
\caption{Pass@K comparison across diverse k list [1,2,4,8,16,32,64,128] with repeated 10 runs on the AIME2024 benchmark with Qwen3-4B-Base as the base model.}
\label{tab:aime24_scale_table}
\end{table*}

\begin{table*}[tbp]
\centering
\resizebox{\textwidth}{!}{%
\begin{tabular}{l *{8}{c}}
\toprule
\textbf{Model} & \textbf{1} & \textbf{2} & \textbf{4} & \textbf{8} & \textbf{16} & \textbf{32} & \textbf{64} & \textbf{128} \\
\midrule
Base         & $7.33 \pm 8.02$ & $10.67 \pm 11.65$ & $15.33 \pm 11.52$ & $20.33 \pm 9.97$ & $26.67 \pm 7.58$ & $30.33 \pm 10.35$ & $37.00 \pm 14.59$ & $44.00 \pm 14.71$ \\
GRPO         & $19.33 \pm 11.49$ & $25.67 \pm 11.00$ & $\mathbf{32.67} \pm 10.97$ & $35.67 \pm 12.86$ & $\mathbf{43.33} \pm 14.16$ & $\mathbf{49.00} \pm 9.92$ & $\mathbf{56.33} \pm 9.02$ & $\mathbf{59.33} \pm 5.49$ \\
DAPO         & $19.33 \pm 9.19$ & $24.00 \pm 7.35$ & $28.00 \pm 11.86$ & $31.33 \pm 10.60$ & $36.33 \pm 10.35$ & $43.00 \pm 6.35$ & $44.33 \pm 5.69$ & $48.00 \pm 4.00$ \\
R++  & $13.33 \pm 11.58$ & $16.00 \pm 8.39$ & $19.00 \pm 7.02$ & $24.00 \pm 8.16$ & $31.33 \pm 9.72$ & $31.00 \pm 9.53$ & $37.00 \pm 6.13$ & $39.33 \pm 4.16$ \\
GSPO         & $12.67 \pm 8.16$ & $17.33 \pm 13.58$ & $20.00 \pm 10.69$ & $26.33 \pm 8.43$ & $28.33 \pm 11.46$ & $35.00 \pm 8.72$ & $39.67 \pm 8.13$ & $44.67 \pm 2.97$ \\
\rowcolor[rgb]{ .867,  .922,  .969} \our & $\mathbf{23.00} \pm 9.97$ & $\mathbf{26.00} \pm 12.60$ & $30.67 \pm 13.63$ & $\mathbf{37.00} \pm 13.39$ & $42.00 \pm 13.79$ & $47.00 \pm 8.49$ & $51.67 \pm 7.67$ & $55.33 \pm 6.83$ \\
\bottomrule

\end{tabular}
}
\caption{Pass@K comparison across diverse k list [1,2,4,8,16,32,64,128] with repeated 10 runs on the AIME2025 benchmark.}
\label{tab:aime25_scale_table}
\end{table*}

\begin{table*}[tbp]
\centering
\resizebox{\textwidth}{!}{%
\begin{tabular}{l *{8}{c}}
\toprule
\textbf{Model} & \textbf{1} & \textbf{2} & \textbf{4} & \textbf{8} & \textbf{16} & \textbf{32} & \textbf{64} & \textbf{128} \\
\midrule
Base         & $32.75 \pm 28.90$ & $46.00 \pm 28.26$ & $58.25 \pm 27.01$ & $71.75 \pm 19.97$ & $81.75 \pm 14.51$ & $87.50 \pm 7.86$ & $91.25 \pm 3.97$ & $95.00 \pm 4.62$ \\
GRPO         & $62.75 \pm 20.46$ & $72.25 \pm 16.61$ & $79.75 \pm 14.44$ & $87.25 \pm 11.54$ & $90.75 \pm 5.27$ & $94.75 \pm 3.25$ & $96.25 \pm 1.97$ & $96.50 \pm 1.22$ \\
DAPO         & $63.50 \pm 15.09$ & $71.25 \pm 14.99$ & $78.75 \pm 10.99$ & $84.00 \pm 9.08$ & $88.00 \pm 6.62$ & $90.50 \pm 3.29$ & $93.50 \pm 2.22$ & $94.00 \pm 1.90$ \\
R++  & $57.50 \pm 20.98$ & $69.75 \pm 19.34$ & $79.00 \pm 13.12$ & $82.75 \pm 9.77$ & $88.50 \pm 6.40$ & $91.75 \pm 3.97$ & $92.00 \pm 2.40$ & $93.50 \pm 2.22$ \\
GSPO         & $57.25 \pm 19.68$ & $72.00 \pm 18.97$ & $77.25 \pm 14.01$ & $82.75 \pm 12.22$ & $89.50 \pm 10.52$ & $94.25 \pm 3.47$ & $95.50 \pm 3.00$ & $97.00 \pm 1.00$ \\
\rowcolor[rgb]{ .867,  .922,  .969} \our & $\mathbf{68.25} \pm 21.69$ & $\mathbf{79.00} \pm 16.98$ & $\mathbf{85.75} \pm 11.72$ & $\mathbf{91.75} \pm 6.37$ & $\mathbf{94.00} \pm 1.90$ & $\mathbf{95.75} \pm 2.37$ & $\mathbf{97.00} \pm 1.00$ & $\mathbf{97.00} \pm 1.00$ \\ \\
\bottomrule

\end{tabular}
}
\caption{Pass@K comparison across diverse k list [1,2,4,8,16,32,64,128] with repeated 10 runs on the AMC23 benchmark with Qwen3-4B-Base as the base model.}
\label{tab:amc23_scale_table}
\end{table*}

\begin{table*}[tbp]
\centering
\resizebox{\textwidth}{!}{%
\begin{tabular}{l *{8}{c}}
\toprule
\textbf{Model} & \textbf{1} & \textbf{2} & \textbf{4} & \textbf{8} & \textbf{16} & \textbf{32} & \textbf{64} & \textbf{128} \\
\midrule
Base         & $0.33 \pm 1.00$ & $1.33 \pm 3.33$ & $3.33 \pm 5.33$ & $6.67 \pm 8.67$ & $8.00 \pm 7.22$ & $11.00 \pm 8.58$ & $14.33 \pm 11.11$ & $20.67 \pm 9.35$ \\
GRPO         & $8.00 \pm 9.19$ & $10.00 \pm 5.63$ & $16.00 \pm 10.60$ & $19.67 \pm 8.27$ & $22.33 \pm 5.67$ & $26.67 \pm 6.30$ & $30.00 \pm 5.27$ & $31.33 \pm 3.06$ \\
DAPO         & $6.67 \pm 8.02$ & $12.67 \pm 9.35$ & $13.67 \pm 6.35$ & $17.00 \pm 6.49$ & $23.67 \pm 6.93$ & $22.00 \pm 4.97$ & $23.67 \pm 5.39$ & $26.67 \pm 3.27$ \\
REINFORCE++  & $4.33 \pm 6.86$ & $6.00 \pm 7.33$ & $11.67 \pm 11.92$ & $16.33 \pm 11.11$ & $21.00 \pm 10.35$ & $27.33 \pm 8.16$ & $32.33 \pm 5.90$ & $33.33 \pm 3.33$ \\
GSPO         & $5.00 \pm 7.39$ & $8.67 \pm 9.06$ & $9.33 \pm 10.53$ & $16.33 \pm 10.59$ & $19.00 \pm 10.43$ & $26.33 \pm 7.00$ & $28.33 \pm 4.33$ & $29.67 \pm 1.00$ \\
\rowcolor[rgb]{ .867,  .922,  .969} \our & $\mathbf{10.00} \pm 8.97$ & $\mathbf{16.33} \pm 6.19$ & $\mathbf{17.33} \pm 9.27$ & $\mathbf{21.67} \pm 13.30$ & $\mathbf{26.67} \pm 14.02$ & $\mathbf{33.67} \pm 14.29$ & $\mathbf{38.00} \pm 12.62$ & $\mathbf{44.00} \pm 7.13$ \\
\bottomrule

\end{tabular}
}
\caption{Pass@K comparison across diverse k list [1,2,4,8,16,32,64,128] with repeated 10 runs on the HMMT25 benchmark with Qwen3-4B-Base as the base model.}
\label{tab:hmmt25_scale_table}
\end{table*}

\begin{table*}[tbp]
\centering
\resizebox{\textwidth}{!}{%
\begin{tabular}{l *{8}{c}}
\toprule
\textbf{Model} & \textbf{1} & \textbf{2} & \textbf{4} & \textbf{8} & \textbf{16} & \textbf{32} & \textbf{64} & \textbf{128} \\
\midrule
Base          & $7.67 \pm 8.93$ & $10.00 \pm 9.19$ & $10.33 \pm 5.30$ & $13.00 \pm 5.97$ & $15.67 \pm 5.53$ & $17.00 \pm 5.30$ & $17.00 \pm 2.33$ & $17.67 \pm 1.53$ \\
GRPO          & $20.00 \pm 9.13$ & $20.33 \pm 10.49$ & $21.33 \pm 6.83$ & $25.33 \pm 5.86$ & $26.67 \pm 2.00$ & $26.33 \pm 3.53$ & $26.67 \pm 0.00$ & $26.67 \pm 0.00$ \\
DAPO          & $22.67 \pm 12.46$ & $19.67 \pm 13.16$ & $23.67 \pm 9.11$ & $25.00 \pm 6.86$ & $25.00 \pm 3.53$ & $27.67 \pm 3.00$ & $27.33 \pm 4.67$ & $28.33 \pm 2.86$ \\
REINFORCE++   & $16.67 \pm 10.16$ & $16.33 \pm 11.39$ & $19.33 \pm 7.19$ & $19.67 \pm 3.30$ & $21.67 \pm 3.67$ & $22.33 \pm 1.53$ & $23.00 \pm 1.00$ & $22.67 \pm 1.33$ \\
GSPO          & $17.00 \pm 14.13$ & $14.67 \pm 8.69$ & $18.00 \pm 7.86$ & $20.33 \pm 5.53$ & $20.33 \pm 3.67$ & $21.00 \pm 1.53$ & $21.00 \pm 1.53$ & $22.00 \pm 2.53$ \\
\rowcolor[rgb]{ .867,  .922,  .969} \our          & $\mathbf{27.67} \pm 15.39$ & $\mathbf{27.33} \pm 15.44$ & $\mathbf{26.33} \pm 6.86$ & $\mathbf{29.33} \pm 8.02$ & $\mathbf{33.67} \pm 7.02$ & $\mathbf{35.00} \pm 3.97$ & $\mathbf{36.00} \pm 3.33$ & $\mathbf{35.67} \pm 2.33$ \\

\bottomrule

\end{tabular}
}
\caption{Maj@K comparison across diverse k list [1,2,4,8,16,32,64,128] with repeated 10 runs on the AIME2024 benchmark with Qwen3-4B-Base as the base model.}
\label{tab:aime24_majk_table}
\end{table*}

\begin{table*}[tbp]
\centering
\resizebox{\textwidth}{!}{%
\begin{tabular}{l *{8}{c}}
\toprule
\textbf{Model} & \textbf{1} & \textbf{2} & \textbf{4} & \textbf{8} & \textbf{16} & \textbf{32} & \textbf{64} & \textbf{128} \\
\midrule

Base          & $6.33 \pm 9.16$ & $6.67 \pm 9.22$ & $8.33 \pm 8.13$ & $12.33 \pm 8.49$ & $12.33 \pm 7.64$ & $15.33 \pm 6.83$ & $18.67 \pm 4.83$ & $19.67 \pm 4.06$ \\
GRPO          & $20.00 \pm 14.02$ & $18.67 \pm 9.86$ & $22.33 \pm 10.79$ & $25.33 \pm 7.65$ & $\mathbf{26.67} \pm 7.57$ & $28.00 \pm 6.11$ & $\mathbf{29.67} \pm 5.83$ & $\mathbf{28.67} \pm 4.16$ \\
DAPO          & $22.33 \pm 9.69$ & $16.00 \pm 8.97$ & $20.00 \pm 8.60$ & $22.67 \pm 6.30$ & $23.33 \pm 6.30$ & $24.00 \pm 5.33$ & $24.67 \pm 3.86$ & $24.67 \pm 3.63$ \\
REINFORCE++   & $12.00 \pm 6.39$ & $13.00 \pm 8.49$ & $16.00 \pm 7.58$ & $16.00 \pm 6.00$ & $17.00 \pm 3.16$ & $18.00 \pm 2.67$ & $16.67 \pm 0.00$ & $16.67 \pm 0.00$ \\
GSPO          & $10.33 \pm 8.79$ & $13.00 \pm 6.49$ & $14.33 \pm 7.16$ & $19.00 \pm 8.86$ & $19.00 \pm 6.49$ & $19.00 \pm 4.63$ & $19.00 \pm 3.67$ & $19.00 \pm 2.63$ \\
\rowcolor[rgb]{ .867,  .922,  .969} \our          & $\mathbf{23.33} \pm 11.86$ & $\mathbf{22.33} \pm 8.33$ & $\mathbf{22.67} \pm 5.19$ & $\mathbf{26.00} \pm 9.72$ & $26.00 \pm 7.72$ & $\mathbf{29.33} \pm 5.49$ & $28.00 \pm 5.27$ & $\mathbf{28.67} \pm 2.67$ \\
\bottomrule

\end{tabular}
}
\caption{Maj@K comparison across diverse k list [1,2,4,8,16,32,64,128] with repeated 10 runs on the AIME2025 benchmark with Qwen3-4B-Base as the base model.}
\label{tab:aime25_majk_table}
\end{table*}

\begin{table*}[tbp]
\centering
\resizebox{\textwidth}{!}{%
\begin{tabular}{l *{8}{c}}
\toprule
\textbf{Model} & \textbf{1} & \textbf{2} & \textbf{4} & \textbf{8} & \textbf{16} & \textbf{32} & \textbf{64} & \textbf{128} \\
\midrule
Base          & $35.00 \pm 31.95$ & $31.50 \pm 26.39$ & $40.00 \pm 24.74$ & $46.00 \pm 16.36$ & $48.75 \pm 14.16$ & $51.00 \pm 14.53$ & $50.00 \pm 12.72$ & $50.75 \pm 9.32$ \\
GRPO          & $62.00 \pm 21.92$ & $62.00 \pm 17.65$ & $66.00 \pm 17.60$ & $66.50 \pm 14.35$ & $69.50 \pm 11.36$ & $69.25 \pm 7.49$ & $70.75 \pm 6.20$ & $70.75 \pm 4.87$ \\
DAPO          & $65.00 \pm 16.14$ & $62.75 \pm 15.39$ & $69.25 \pm 13.22$ & $69.00 \pm 11.91$ & $70.25 \pm 9.89$ & $73.75 \pm 5.10$ & $73.25 \pm 4.65$ & $73.75 \pm 4.15$ \\
REINFORCE++   & $57.75 \pm 20.72$ & $58.75 \pm 21.67$ & $63.75 \pm 16.72$ & $67.00 \pm 12.09$ & $70.00 \pm 9.24$ & $72.50 \pm 8.35$ & $73.25 \pm 7.54$ & $74.25 \pm 6.04$ \\
GSPO          & $56.75 \pm 22.36$ & $60.00 \pm 21.14$ & $64.75 \pm 18.44$ & $65.50 \pm 15.02$ & $69.25 \pm 12.06$ & $73.00 \pm 9.45$ & $73.00 \pm 6.02$ & $74.50 \pm 3.50$ \\
\rowcolor[rgb]{ .867,  .922,  .969} \our          & $\mathbf{68.50} \pm 22.07$ & $\mathbf{70.00} \pm 17.77$ & $\mathbf{76.25} \pm 17.77$ & $\mathbf{84.25} \pm 10.27$ & $\mathbf{84.75} \pm 7.04$ & $\mathbf{84.50} \pm 4.90$ & $\mathbf{85.00} \pm 3.90$ & $\mathbf{85.50} \pm 1.90$ \\
\bottomrule

\end{tabular}
}
\caption{Maj@K comparison across diverse k list [1,2,4,8,16,32,64,128] with repeated 10 runs on the AMC23 benchmark with Qwen3-4B-Base as the base model.}
\label{tab:amc23_majk_table}
\end{table*}

\begin{table*}[tbp]
\centering
\resizebox{\textwidth}{!}{%
\begin{tabular}{l *{8}{c}}
\toprule
\textbf{Model} & \textbf{1} & \textbf{2} & \textbf{4} & \textbf{8} & \textbf{16} & \textbf{32} & \textbf{64} & \textbf{128} \\
\midrule
Base          & $1.00 \pm 3.00$ & $1.67 \pm 2.63$ & $0.67 \pm 1.33$ & $2.67 \pm 1.33$ & $2.33 \pm 1.53$ & $3.00 \pm 1.00$ & $3.00 \pm 1.00$ & $3.33 \pm 0.00$ \\
GRPO          & $6.67 \pm 5.33$ & $8.67 \pm 9.27$ & $10.33 \pm 5.67$ & $10.67 \pm 6.02$ & $10.00 \pm 2.00$ & $12.67 \pm 2.53$ & $13.00 \pm 1.00$ & $12.67 \pm 1.33$ \\
DAPO          & $8.00 \pm 7.33$ & $6.33 \pm 8.93$ & $8.33 \pm 6.72$ & $9.67 \pm 3.30$ & $10.33 \pm 3.30$ & $11.67 \pm 2.63$ & $12.00 \pm 2.53$ & $12.67 \pm 1.33$ \\
REINFORCE++   & $5.67 \pm 9.00$ & $3.33 \pm 4.53$ & $5.67 \pm 4.63$ & $6.00 \pm 2.97$ & $6.33 \pm 4.63$ & $5.67 \pm 2.63$ & $5.33 \pm 2.97$ & $5.67 \pm 3.16$ \\
GSPO          & $4.00 \pm 7.49$ & $6.00 \pm 8.60$ & $5.00 \pm 3.67$ & $6.33 \pm 2.86$ & $8.33 \pm 4.86$ & $9.00 \pm 3.00$ & $8.67 \pm 1.63$ & $8.67 \pm 1.63$ \\
\rowcolor[rgb]{ .867,  .922,  .969} \our          & $\mathbf{11.00} \pm 8.49$ & $\mathbf{11.00} \pm 4.79$ & $\mathbf{11.67} \pm 5.69$ & $\mathbf{12.67} \pm 5.79$ & $\mathbf{15.33} \pm 3.86$ & $\mathbf{17.67} \pm 2.33$ & $\mathbf{16.67} \pm 0.00$ & $\mathbf{16.67} \pm 0.00$ \\
\bottomrule

\end{tabular}
}
\caption{Maj@K comparison across diverse k list [1,2,4,8,16,32,64,128] with repeated 10 runs on the HMMT25 benchmark with Qwen3-4B-Base as the base model.}
\label{tab:hmmt25_majk_table}
\end{table*}

\begin{table*}[tbp]
  \centering
         \resizebox{\linewidth}{!}{%
    \begin{tabular}{lcccccccccccccc}
    \toprule
    \multirow{2}[4]{*}{\textbf{Model}} & \multicolumn{2}{c}{\textbf{AIME2024}} & \multicolumn{2}{c}{\textbf{AIME2025}} & \multicolumn{2}{c}{\textbf{AMC23}} & \multicolumn{2}{c}{\textbf{HMMT25}} & \multicolumn{2}{c}{\textbf{MATH}} & \multicolumn{2}{c}{\textbf{OlympiadB}} & \multicolumn{2}{c}{\textbf{Avg}} \\
\cmidrule(r){2-3} \cmidrule(r){4-5} \cmidrule(r){6-7} \cmidrule(r){8-9} \cmidrule(r){10-11} \cmidrule(r){12-13} \cmidrule(r){14-15}          & \textbf{P@1} & \textbf{P@K} & \textbf{P@1} & \textbf{P@K} & \textbf{P@1} & \textbf{P@K} & \textbf{P@1} & \textbf{P@K} & \textbf{P@1} & \textbf{P@K} & \textbf{P@1} & \textbf{P@K} & \textbf{P@1} & \textbf{P@K} \\
    \midrule
    Base  & 9.51  & 50.00 & 6.65  & {43.33} & 32.91 & 97.50 & 0.99  & 26.67 & 58.89 & 91.00 & 31.81 & 65.40 & 23.46 & 62.32 \\
    BAPO  & 17.86 & 50.00 & 16.69 & 56.67 & 61.04 & 95.00 & 8.57  & 33.33 & 82.81 & 94.40 & 51.32 & 70.40 & 39.72 & 66.63 \\
    KL-COv & 17.73 & 53.33 & 20.26 & 40.00 & 59.96 & 90.00 & 7.16  & 23.33 & 83.38 & 93.20 & 49.63 & 68.85 & 39.69 & 61.45 \\
    Clip-Cov & 15.55 & 60.00 & 14.40 & 46.67 & 59.98 & 97.50 & 4.48  & 36.67 & 80.98 & 94.40 & 49.02 & 71.94 & 37.40 & 67.86 \\
    Entropy-Adv & 17.71 & 63.33 & 15.99 & 50.00 & 61.86 & 97.50 & 6.51  & 30.00 & 81.73 & 94.80 & 50.17 & 73.67 & 38.99 & 68.22 \\
    \rowcolor[rgb]{ .867,  .922,  .969} \our  & \textbf{25.86} & \textbf{73.33} & \textbf{22.97} & \textbf{60.00} & \textbf{69.65} & \textbf{97.50} & \textbf{10.81} & \textbf{46.67} & \textbf{86.93} & \textbf{95.80} & \textbf{57.07} & \textbf{76.25} & \textbf{45.55} & \textbf{74.92} \\
    \bottomrule
    \end{tabular}%
    }
      \caption{Comparison with other exploration-enhanced algorithms. Given numerous hyperparameter combinations, these methods show bad generalization with their officially suggested hyperparameters.}
  \label{tab:cmp_other_entropy}%
\end{table*}%

\end{document}